\documentclass[11pt]{article}

\usepackage[preprint]{acl}
\usepackage{times}
\usepackage{latexsym}
\usepackage[T1]{fontenc}
\usepackage[utf8]{inputenc}
\usepackage{microtype}
\usepackage{inconsolata}
\usepackage{graphicx}
\usepackage{enumitem}
\usepackage{amsmath}
\usepackage{hyperref}       
\usepackage{url}            
\usepackage{booktabs}       
\usepackage{amsfonts}       
\usepackage{nicefrac}       
\usepackage{xcolor}         
\usepackage{subcaption}
\usepackage{amssymb}   
\usepackage{bbm}       
\usepackage{xspace} 
\usepackage[table]{xcolor}
\usepackage{booktabs} 

\newbool{showcomments}
\booltrue{showcomments}
\newcommand{\cmt}[3]{%
    \ifbool{showcomments}{%
        {\color{#1}\textbf{#2}: #3}%
    }{}%
}

\newcommand{\model}{{AdaSTORM}\xspace}
\title{AdaSTORM: Scaling LLM Reasoning on Dynamic Graphs via Adaptive Spatio-Temporal Multi-Agent Collaboration}

\author{
Bing Hao\textsuperscript{1}\thanks{Equal contribution.}, Ruijie Wang\textsuperscript{2}\footnotemark[1], Haodong Qian\textsuperscript{1}\footnotemark[1],
Yunlong Chu\textsuperscript{1},\\
\textbf{Yuhang Liu\textsuperscript{1}},
\textbf{Yumeng Lin\textsuperscript{1}},
\textbf{Minglai Shao\textsuperscript{1}\thanks{Corresponding author.}},
\textbf{Jianxin Li\textsuperscript{2}}
\\[0.5em]
\textsuperscript{1}Tianjin University, China \qquad
\textsuperscript{2}Beihang University, China
\\[0.5em]
\texttt{\{haobing, qianhd, cyl2024245030, liuyuhang\_13, lym619, shaoml\}@tju.edu.cn}
\\
\texttt{\{ruijiew, lijx\}@buaa.edu.cn}
}

\begin{document}
\maketitle
\begin{abstract}
Large Language Models (LLMs) demonstrate remarkable potential in dynamic graph reasoning, but suffer from a scaling bottleneck: current models can only handle graphs with tens of nodes, constrained by exponential reasoning overhead and finite context windows. While multi-agent systems (MAS) offer collective reasoning and topology-aware orchestration, capabilities naturally suited for graph-structured tasks, their application to dynamic graphs remains unexplored. This paper presents Scaling LLM Reasoning on Dynamic Graphs via Adaptive Spatio-Temporal Multi-Agent Collaboration (AdaSTORM), a framework that reformulates large-scale dynamic graph reasoning into two stages: (i) \textbf{Adaptive Partitioning}, partitioning large-scale dynamic graphs into subregions that match the model's reasoning capacity while minimizing inference cost; and (ii) \textbf{Collaborative Reasoning}, aligning graph partition topologies with a spatio-temporal decoupled multi-agent architecture. AdaSTORM is the first multi-agent framework tailored for dynamic graph reasoning. Extensive experiments show that AdaSTORM successfully breaks through the scaling bottleneck, scaling reasoning to thousand-node graphs with over 90\% accuracy across several large-scale dynamic graph settings without external tools, significantly outperforms seven competitive 
baselines. Furthermore, it achieves state-of-the-art accuracy on 
existing benchmarks and generalizes robustly to real-world datasets. The source code is available at: \url{https://github.com/irisorchid107/AdaSTORM/}.

\end{abstract}


\begin{figure}[t] 
    \centering
    \includegraphics[width=1.0\linewidth]{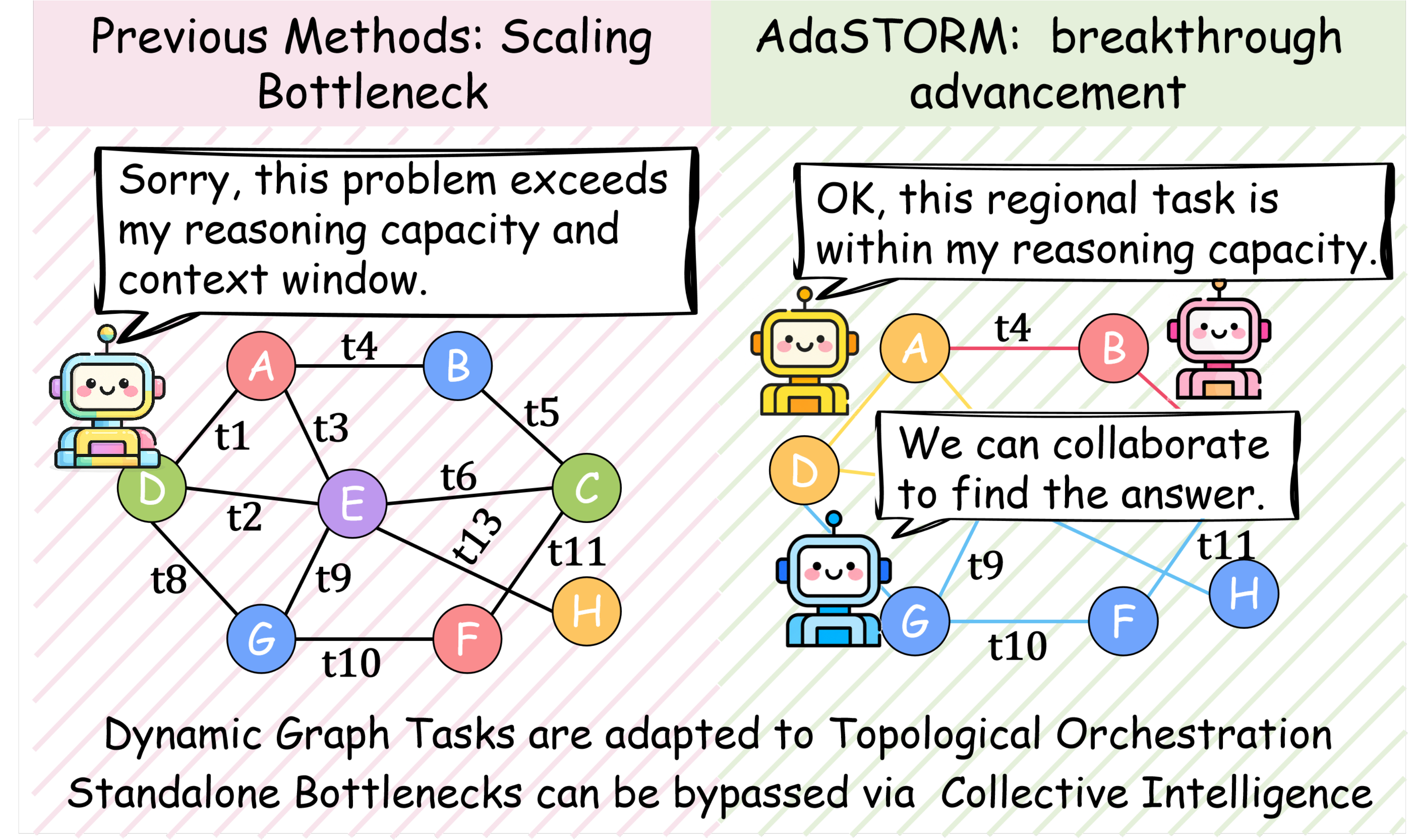} 
    \caption{The core motivation of AdaSTORM. Left: Standalone LLMs 
suffer from severe reasoning capacity and contextual window constraints 
on large-scale dynamic graphs. Right: AdaSTORM bypasses these bottlenecks through collective reasoning.}
    \label{fig:motivation}
    \vspace{-3mm}
\end{figure}

\section{Introduction}
Large Language Models (LLMs) show promise in reasoning over 
graph-structured data~\cite{mao2024advancinggraphrepresentationlearning,jin2024largelanguagemodelsgraphs}, with dynamic graph reasoning standing out as a critical yet 
challenging frontier~\cite{zhang2024llm4dyglargelanguagemodels}, as it captures 
the temporal evolution of relational data in applications ranging from social networks to traffic forecasting~\cite{10.1145/3627673.3679602,10.1609/aaai.v38i13.29381,10.1145/3627673.3679675}. 

\begin{figure} 
    \centering
    \includegraphics[width=1.0\linewidth]{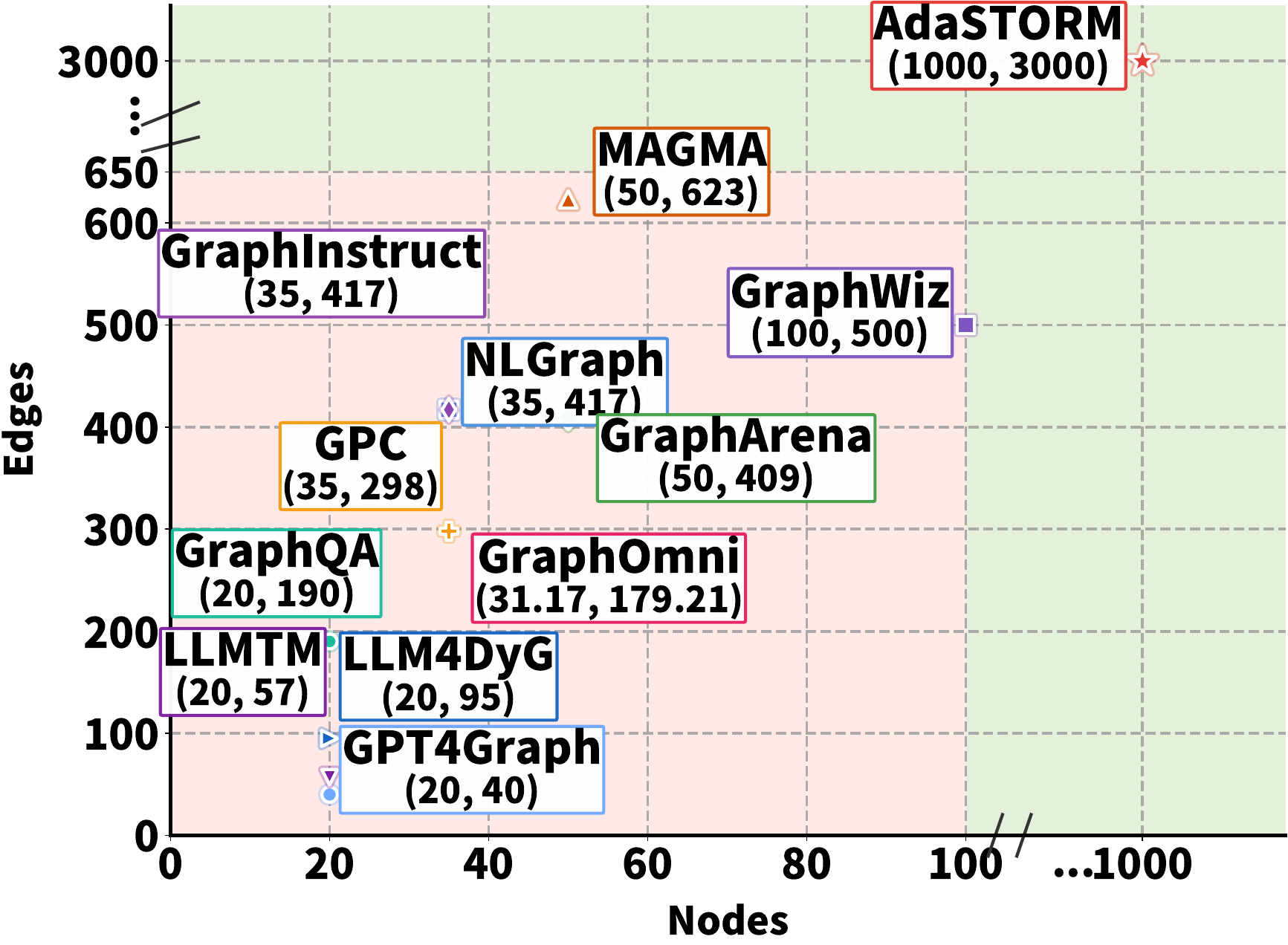} 
    \caption{Comparison of graph scales handled by existing benchmarks 
and methods versus \model{}. ($\#nodes$, $\#edges$) reflects the maximum  node/edge counts reported in the respective papers.}
    \label{fig:benchmarks}
    \vspace{-3mm}
\end{figure}

Recent literature has explored two primary trajectories: benchmark evaluation~\cite{hao2026llmtmbenchmarkingoptimizingllms,dai2025largelanguagemodelsunderstand,xu2026graphomnicomprehensiveextensiblebenchmark,tang2025grapharenaevaluatingexploringlarge} and methodological development~\cite{11222956,chen2024graphwizinstructionfollowinglanguagemodel,luo2025graphinstructempoweringlargelanguage}. Despite these advancements, current methods remain confined to small-to-medium scale graphs (typically tens of nodes), as illustrated in Fig.~\ref{fig:benchmarks}. This limitation stems from a fundamental scaling bottleneck: as graph size increases, structural complexity rapidly exceeds the reasoning capacity and context window of a standalone model, inducing severe hallucinations and catastrophic performance collapse~\cite{tang2025grapharenaevaluatingexploringlarge}.

Multi-Agent Systems (MAS)~\cite{chen2025surveyllmbasedmultiagentsystem,guo2024largelanguagemodelbased} overcome the reasoning capacity and contextual window limits of standalone LLMs by decomposing complex problems into collaborative sub-tasks~\cite{wu2026talkrightspecialistsiterative,hong2024metagptmetaprogrammingmultiagent,wu2023autogenenablingnextgenllm}. By orchestrating agents through structured topologies~\cite{qian2024chatdevcommunicativeagentssoftware, du2023improvingfactualityreasoninglanguage}, the communication architecture of MAS inherently mirrors the structural characteristics of graphs, rendering MAS naturally compatible with dynamic graph reasoning (Fig.~\ref{fig:motivation}). However, standard multi-agent paradigms (e.g., chain- and debate- based) fail to handle the combinatorial explosion triggered by graph scale~\cite{Lu2025KARMALM,fourney2024magenticonegeneralistmultiagentsolving,xu2025languageagentsreinforcementlearning}, as they typically operate globally on the entire graph topology while lacking mechanisms for large-scale graph partitioning. To bridge this gap, we present \textbf{AdaSTORM}, a multi-agent framework that reformulates large-scale dynamic graph reasoning by exploiting the inherent partitionability~\cite{patwary2021sdpscalablerealtimedynamic,10.1145/3357384.3358088} and spatio-temporal decoupling~\cite{zhang2024llm4dyglargelanguagemodels} of dynamic graphs.

Specifically, AdaSTORM operates in two sequential stages: (i) \textbf{Adaptive Partitioning} tailors graph decomposition to model capacity. A capacity estimator assesses model feasibility by fusing learnable latent vectors with multimodal features, including LM-encoded queries, model profile, and GNN-based structural encodings. Guided by this capacity estimator and a cost simulator, a reinforcement learning (RL)-based adaptive partitioner learns a policy network to recursively split subregions and migrate nodes, deriving an optimal graph partition that ensures each subregion match the model's reasoning capacity while minimizing inference cost; (ii) \textbf{Collaborative Reasoning} aligns graph partition topologies with a spatio-temporal decoupled multi-agent architecture. By deploying specialized spatial and temporal agents for parallel localized inference, and leveraging cooperative message passing across boundary cut-edges for collective 
reasoning, the framework substantially boosts both the accuracy and scale of dynamic graph tasks for LLMs without external tools.

Our contributions are summarized as follows:

\noindent\textbf{Framework}. AdaSTORM is the first multi-agent framework tailored for dynamic graph reasoning. By exploiting the inherent partitionability, spatio-temporal decoupling and topological affinity of dynamic graphs with multi-agent orchestration, our framework effectively bypasses the scaling bottleneck of standalone LLMs, substantially boosting both the accuracy and scale of dynamic graph tasks for LLMs without external tools.

    
\noindent\textbf{Partitioning}. We design a reasoning capacity-guided reinforcement learning scheme that adaptively partitions large-scale graphs. Driven by flexible reward functions and custom adaptation criteria, this objective-agnostic framework seamlessly generalizes to diverse partitioning goals.

\noindent\textbf{Experiments}. Extensive evaluation validates that AdaSTORM effectively breaks througt the scaling bottleneck, consistently outperforms single-agent and multi-agent baselines across graph scales. Specifically, it achieves state-of-the-art (SOTA) performance against existing small-to-medium scale baselines (Fig.~\ref{fig:radar}). Notably, it maintains over 90\% accuracy on several large-scale dynamic graph reasoning settings (Table~\ref{table:main}) and generalizes effectively to real-world datasets.

\section{Preliminaries and Problem Formulation}
\textbf{Dynamic Graph}. A dynamic graph is defined as $\mathcal{G} = (\mathcal{V}, \mathcal{E})$, where $\mathcal{V}$ is the node set and $\mathcal{E}$ is a time-ordered sequence of events. To explicitly capture richer structural dynamics, we represent each event as a lifecycle quadruplet $(u, v, s, t) \in \mathcal{E}$, where $s$ and $t$ 
denote the emergence and deletion timestamps, respectively. The static projection of $\mathcal{G}$ is denoted as $G = (\mathcal{V}, E_s)$ 
with $E_s = \{ \{u, v\} \mid \exists (u, v, s, t) \in \mathcal{E} \}$. More Preliminaries are deferred to Appendix~\ref{app:preliminaries}.




\noindent\textbf{Reasoning Feasibility}. Given an LLM $\mathcal{M}$ and a reasoning task $\mathcal{T}$, we define the reasoning feasibility of $\mathcal{M}$ on a graph region $P_i$ as the probability of successfully outputting the ground-truth answer $y^\star$:
\begin{equation}
\phi_i \equiv \phi_{\mathcal{M}}(P_i, \mathcal{T}) = \Pr[\mathcal{M}(P_i, \mathcal{T}) = y^\star].
\end{equation}
A subregion $P_i$ is reasoning-feasible if $\phi_i \geq \tau$, where $\tau$ is a predefined performance threshold. This metric quantifies the model's inherent reasoning boundaries on structured data.

\noindent\textbf{Inference Cost}. Given a partitioning $\mathcal{P}$ and a task $\mathcal{T}$, the inference cost $C(\mathcal{P})$ is the total number of agent invocations required to solve the task.

\noindent\textbf{Problem}. Given a dynamic graph $\mathcal{G}$, an LLM $\mathcal{M}$, and a reasoning task $\mathcal{T}$, our goal is to perform inference via a multi-agent architecture, formulated as a two-stage sequential process:

(i) Adaptive Partitioning: Derive an optimal partitioning $\mathcal{P}$ that minimizes inference cost $C(\mathcal{P})$ while bounded by the model's 
reasoning capacity:
\begin{equation}
     \min_{\mathcal{P}} C(\mathcal{P}) \quad \text{s.t.} \quad \phi_i \geq \tau, \quad \forall P_i \in \mathcal{P}.
\end{equation}

(ii) Collaborative Reasoning: Given $\mathcal{P}$, resolve complex dynamic 
graph reasoning tasks via multi-agent collaboration, maximizing both accuracy 
and efficiency.


\section{Methodology}
\begin{figure*}
    \centering
    \includegraphics[width=\linewidth]{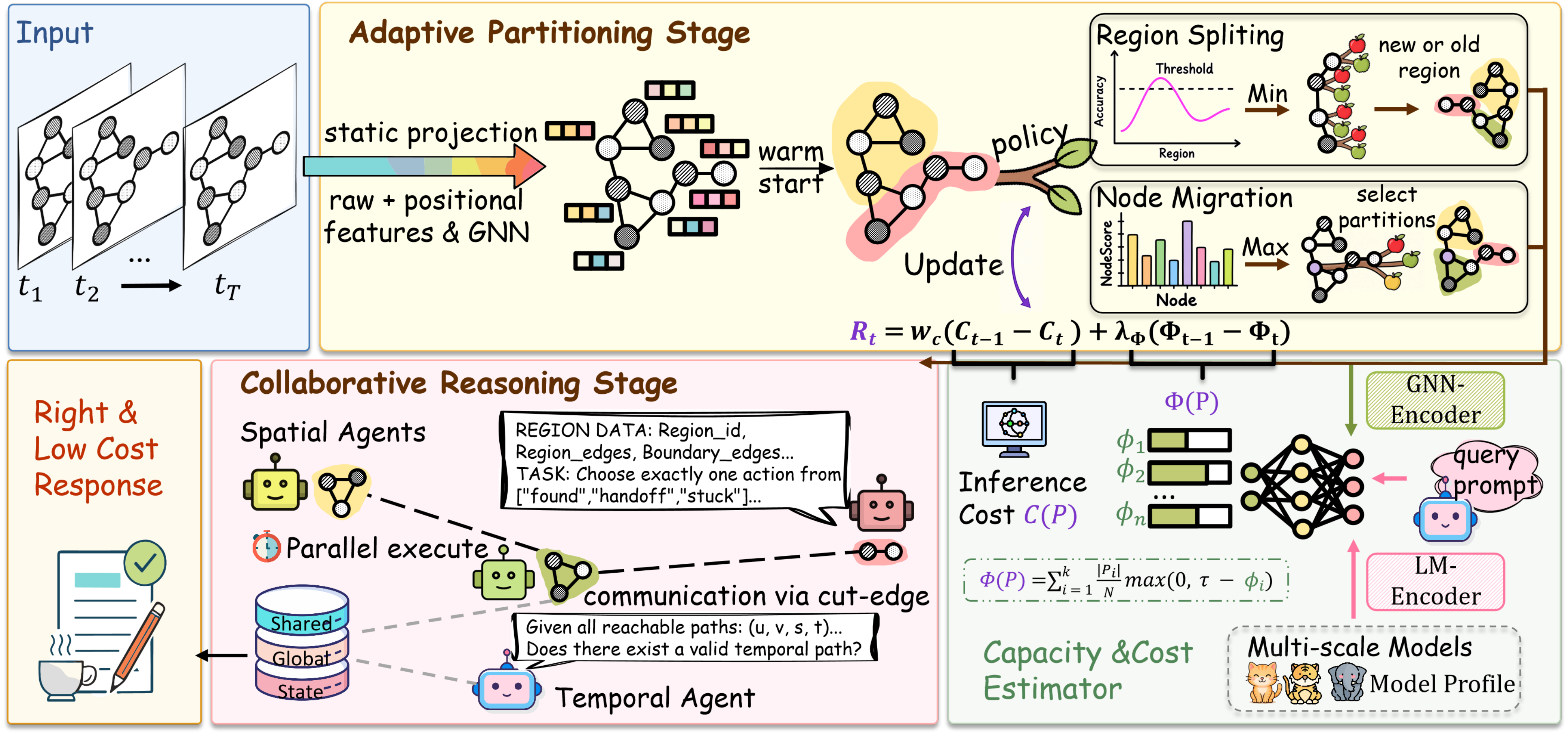}
    \caption{\textbf{Overall framework of AdaSTORM.} In Adaptive Partitioning Stage, we partition large-scale dynamic graphs into capacity-matched, cost-minimizing subregions. In Collaborative Reasoning Stage, we align partition topologies with spatio-temporally decoupled multi-agent orchestration for parallel localized inference, coordinating 
across cut-edges to achieve accurate and cost-effective responses without external tools.}
    \label{fig:framework}
\end{figure*}

Figure~\ref{fig:framework} illustrates the overall architecture of AdaSTORM, which consists of two sequential stages: (i) Adaptive Partitioning tailors graph decomposition to model capacities, where a capacity constrainer and a cost simulator co-guide an RL-based adaptive partitioner, thereby partitioning large-scale graphs into localized subregions, ensuring each subregion matches the model's reasoning capacity while minimize inference cost; (ii) Collaborative Reasoning aligns partition topologies with multi-agent orchestration, deploying specialized spatial and temporal agents for parallel localized inference and coordinating across cut-edges to achieve accurate and 
cost-effective responses without external tools.


\subsection{ Adaptive Partitioning Stage}
\subsubsection{Capacity Estimator}
To guide adaptive partitioner, the Capacity Estimator predicts the feasibility metric $\phi_i$, which quantifies the probability that a target model successfully resolves a given query over subregion $P_i$.

This module bridges semantic reasoning with graph topology by constructing a multimodal joint representation $\mathbf{z}_i$. Specifically, it fuses four orthogonal modalities: LM-encoded task queries; LM-encoded model profiles synthesized 
via an auxiliary LLM to characterize high-level semantic attributes, including model capabilities and usage scenarios (see Appendix~\ref{app:model_profiles}); GNN-based topological encodings that uncover intrinsic structural properties; and a learnable latent vector designed to capture fine-grained behavioral nuances and adaptive task signals that transcend explicit descriptors.



Based on the joint embedding $\mathbf{z}_i$, a Multi-Layer Perceptron ($\text{MLP}_{\text{est}}$) 
estimates the regional feasibility. The network is optimized via a binary cross-entropy criterion:
\begin{equation}
\label{eq:cognitive_estimator}
\begin{split}
\hat{\phi}_i &= \text{MLP}_{\text{est}}(\mathbf{z}_i), \\
\mathcal{L}_{\text{est}} &= -y_i \log(\hat{\phi}_i) - (1 - y_i) \log(1 - \hat{\phi}_i),
\end{split}
\end{equation}
where $y_i \in \{0, 1\}$ is the ground-truth verification label indicating 
whether the target model correctly resolves the task over subregion $P_i$.




\subsubsection{Cost Estimator}
To guide adaptive partitioning, we employ a Cost Estimator to evaluate the inference cost of a given partitioning strategy. Specifically, given a partitioning $\mathcal{P}$ and a task $\mathcal{T}$, the estimator programmatically 
traces the multi-agent execution process and returns the inference cost $C(\mathcal{P})$, 
which is defined as the total number of agent invocations required to resolve the task.

\subsubsection{RL-based Adaptive Partitioner}
\begin{figure}[t]
    \centering
    \begin{subfigure}{0.46\linewidth}
        \centering
        \includegraphics[width=\linewidth]{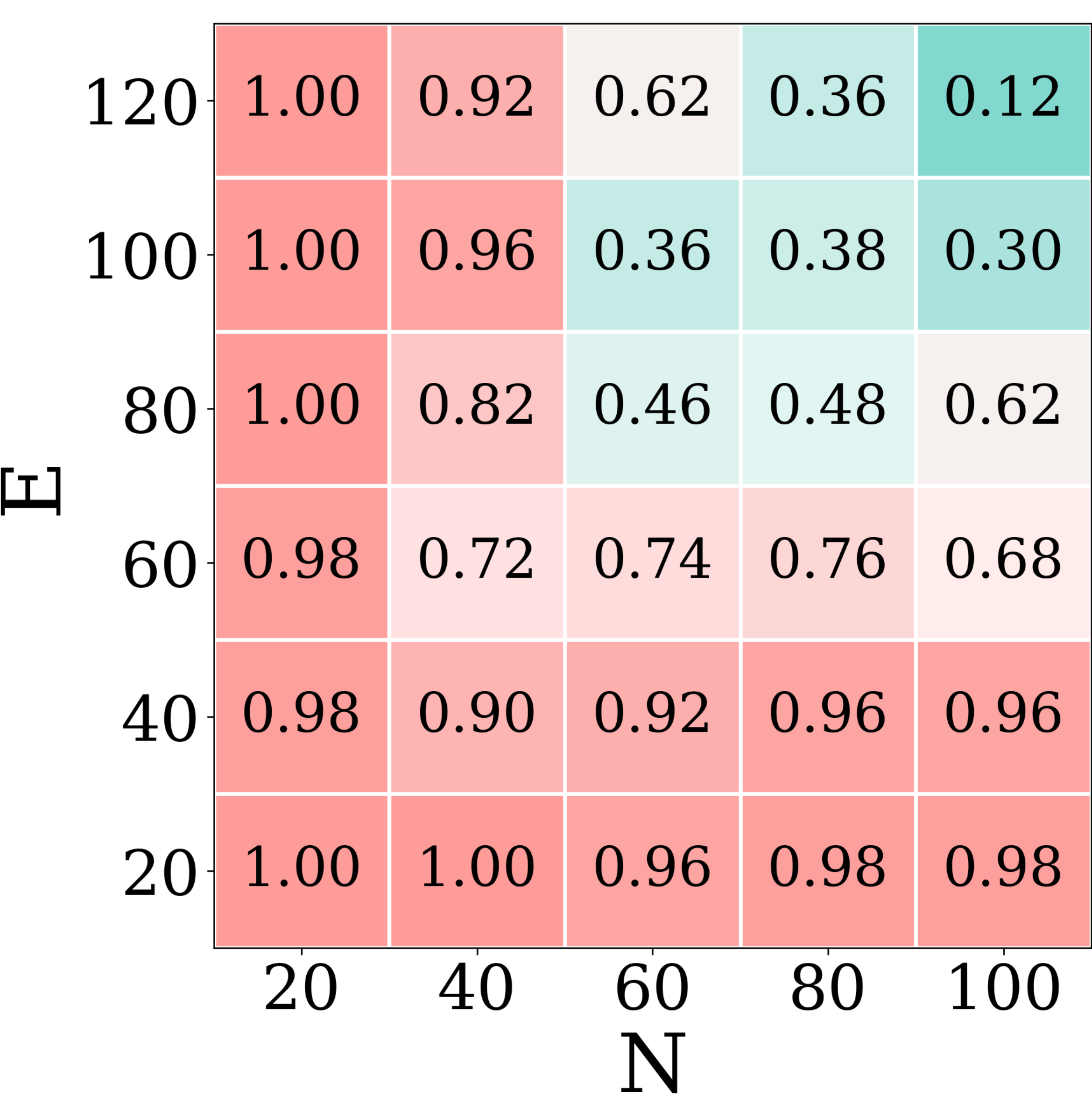}
        \caption{Connected Components}
        \label{fig:heatmap_a}
    \end{subfigure}
    \hspace{0.0mm}
    \begin{subfigure}{0.49\linewidth}
        \centering
        \includegraphics[width=\linewidth]{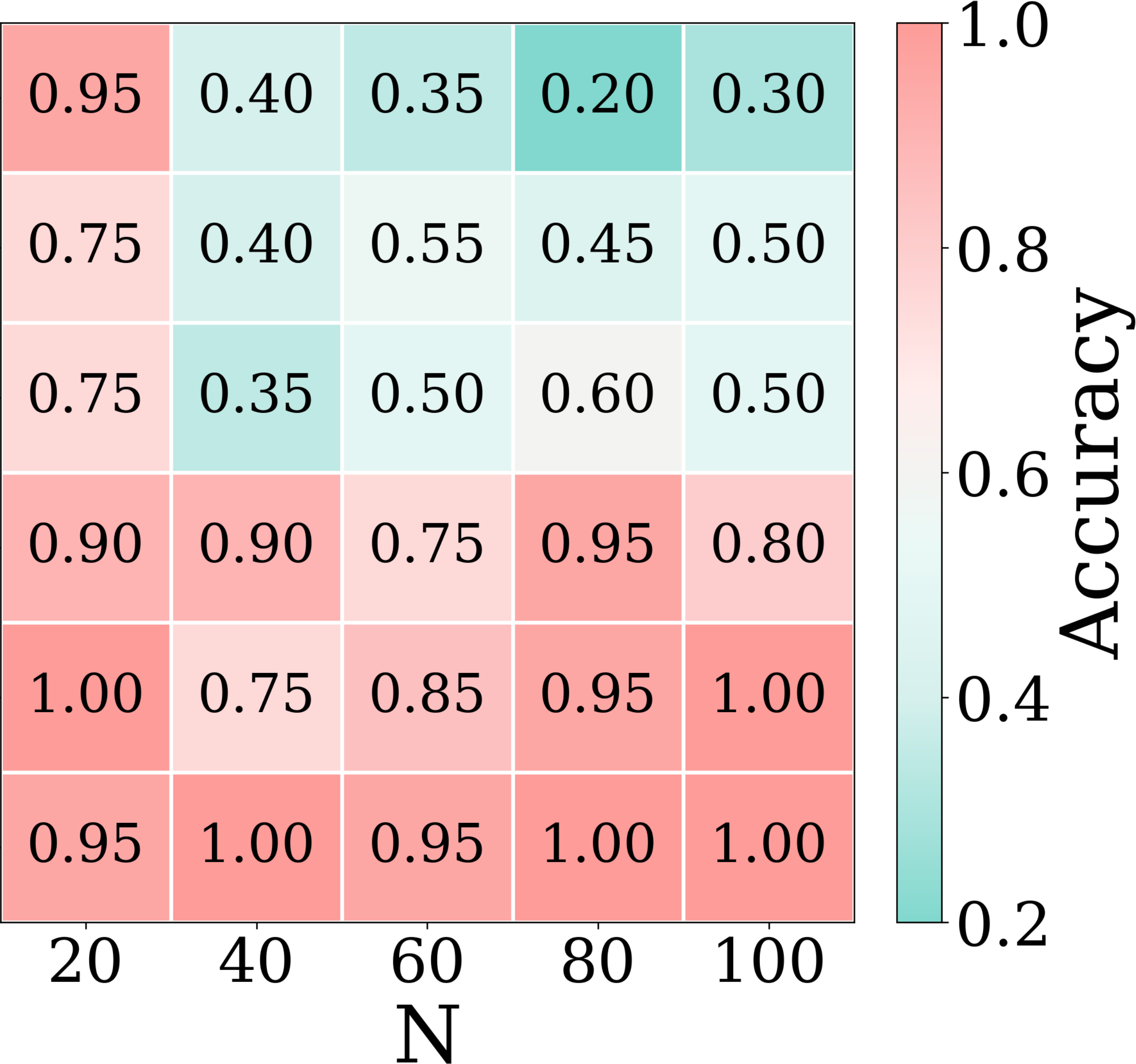}
        \caption{Community Detection}
        \label{fig:heatmap_b}
    \end{subfigure}  \caption{Reasoning performance of the identical model across distinct topological tasks under varying node ($N$) and edge ($E$) scales. Extended results on remaining tasks in~\ref{app:extended_profiles}.}
    \label{fig:heatmap}
    \vspace{-3mm}
\end{figure}

Existing graph partitioning paradigms typically rely on a pre-defined number of partitions $k$~\cite{nazi2019gapgeneralizableapproximategraph,inproceedings,10.5555/2968826.2969013,Tsourakakis2014FENNELSG,10.1145/3097983.3098033}. Yet, a model's reasoning capacity for a given task cannot be  easily quantified via fixed node or edge counts. Moreover, as evidenced in Figure~\ref{fig:heatmap}, the reasoning performance of the identical model varies drastically across diverse tasks, rendering a universal $k$ sub-optimal. To address this, we propose the Adaptive Partitioner, which dynamically decomposes the graph into $k$ optimal partitions, where for the given task, each partition perfectly matches the model's reasoning capacity while minimizing inference cost.
Given an input dynamic graph $\mathcal{G}$, we first derive its static projection $G$ and initialize the partitioning of $G$ into $k_0$ regions via a warm-start graph partitioner, providing a structured starting point for subsequent refinement. 

For the current partitioning $\mathcal{P}$, The pre-trained Capacity Estimator evaluates the feasibility $\phi_i$ of each region $P_i$ under the target model--query pair. To quantify the global violation of reasoning feasibility, we define 
the global infeasibility gap as:
\begin{equation}
\Phi(\mathcal{P})
=
\sum_{i=1}^{k}
\frac{|P_i|}{N}
\max(0,\tau-\phi_i),
\label{eq:global_infeasibility_gap}
\end{equation}
where $N$ is the total number of nodes, $|P_i|$ denotes the size of the $i$-th partition and $\tau$ is the feasibility threshold. Here, a higher $\tau$ imposes a stricter feasibility constraint, while a smaller $\Phi(\mathcal{P})$ indicates that more regions are within the model's reasoning capacity.
Concurrently, the Cost Estimator computes the inference cost $C(\mathcal{P})$ by programmatically tracing the multi-agent execution process.

The Adaptive Partitioner refines $\mathcal{P}$ with two complementary actions. 
\textit{Region splitting} subdivides infeasible regions to reduce the global infeasibility gap, where each node within the target region is either assigned to the new subregion or kept in the old one. \textit{node migration} moves misplaced boundary nodes across partitions to reduce the inference cost. 
At each step, the policy selects one refinement action conditioned on the current feasibility and inference cost.
The reward is defined by the improvement across these dual objectives:
\begin{equation}
\begin{split}
r_t = \, 
&w_c(C_{t-1}-C_t)
+
\lambda_\Phi(\Phi_{t-1}-\Phi_t) \\
&-
\lambda_{\mathrm{split}}\mathbbm{1}_{\mathrm{split}}
-
\lambda_{\mathrm{reject}}\mathbbm{1}_{\mathrm{reject}},
\end{split}
\label{eq:step_reward}
\end{equation}
where $C_t$ is the inference cost, $\Phi_t$ is the global infeasibility gap (Eq.~\eqref{eq:global_infeasibility_gap}), $\mathbbm{1}_{\mathrm{split}}$ and $\mathbbm{1}_{\mathrm{reject}}$ penalize 
unnecessary over-segmentation and ineffective refinements, respectively. 

We train the policy with policy-gradient optimization. At inference time, the learned policy progressively refines the warm-start partitioning into feasibility-satisfying and inference-efficient subregions. The full MDP formulation and training details in Appendix~\ref{app:adaptive_partitioner}.


The adoption of reinforcement learning is motivated by three core insights. First, graph partitioning is NP-hard, making optimal supervision computationally intractable. Second, Our optimization objective is non-differentiable, as it depends on discrete reasoning feasibility and inference cost. Third, Partition refinement is inherently sequential: each split or migration affects later decisions. We therefore train the partitioner to maximize the cumulative reward over the refinement trajectory.


\subsection{Collaborative Reasoning Stage}

Existing literature indicates that prompt-level spatio-temporal decoupling effectively enhances dynamic graphs reasoning in single LLMs~\cite{zhang2024llm4dyglargelanguagemodels}. Therefore, after adaptive partitioning, we aligns the resulting partition topologies with a spatio-temporal decoupled multi-agent architecture for collaborative reasoning. Specifically, Spatial Agents are assigned to subregions to handle static local topology reasoning and cross-region communication, while Temporal Agents operate on dynamic events to enforce chronological constraints. LangGraph~\cite{wang2024agentailanggraphmodular} maintains a shared global state, enabling independent agents to execute in parallel, exchange intermediate results, and update their outputs to achieve accurate and efficient responses without external tools.



\textbf{Spatial Agent}. A Spatial Agent is assigned to each subregion to handle static local topology reasoning. Each agent ingests a local subgraph defined by its nodes, edges, and boundary metadata (Appendix~\ref{app:prompt_design}). Collaboration is managed via the LangGraph-maintained shared global state: updates and messages are committed to this central repository, allowing a dispatcher to route subsequent subregion tasks based on target identifiers.

\textbf{Temporal Agent}. The Temporal Agent operates on dynamic events mapped from static edges (Appendix~\ref{app:preliminaries}) or the original dynamic graph to enforce chronological constraints, focusing on the occurrence, recurrence, and persistence of events (Appendix~\ref{app:prompt_design}). 
Driven by query requirements rather than a rigid pipeline, the Temporal Agent can acts 
as either a pre-filter for search-space pruning or a post-validator to refine spatial reasoning outcomes.

The LangGraph shared global state enables parallel execution of independent agents. By jointly updating outputs from both Spatial and Temporal Agents, this orchestration achieve accurate and efficient responses without external tools.

\section{Experiments}
We evaluate AdaSTORM through three research questions:

\textbf{RQ1:} How does AdaSTORM perform on large-scale dynamic graph tasks, particularly in comparison to leading closed-source models such as DeepSeek-V4-Flash?

\textbf{RQ2:} How does AdaSTORM perform on existing small-to-medium scale graph reasoning benchmarks?

\textbf{RQ3:} How well does AdaSTORM generalize to diverse real-world datasets?

In addition, we conduct comprehensive ablation studies to isolate and verify the contributions of each component, including the Capacity Estimator, the Adaptive Partitioner, the dynamic graph quadruple representation, spatio-temporal decoupling, and multi-agent coordination. 

\subsection{Experimental Setup}
\textbf{Models and Baselines}. We backbone AdaSTORM with DeepSeek-R1-Distill-Qwen-7B, 
14B, and 32B~\cite{deepseekai2025deepseekr1incentivizingreasoningcapability}, denoted as Ours-DQwen-7B, Ours-DQwen-14B, and Ours-DQwen-32B, respectively. Baseline comprise: (i) standalone LLMs, including 
DeepSeek-V4-Flash~\cite{deepseekai2026deepseekv4}, GPT-4o mini~\cite{openai2024gpt4ocard}, 
and vanilla DeepSeek-R1-Distill-Qwen backbones; (ii) chain-based~\cite{qian2024chatdevcommunicativeagentssoftware} 
and debate-based~\cite{du2023improvingfactualityreasoninglanguage} multi-agent 
architectures; and (iii) graph-reasoning specialized frameworks, including GraphWiz~\cite{chen2024graphwizinstructionfollowinglanguagemodel}, 
GraphArena~\cite{tang2025grapharenaevaluatingexploringlarge}, GraphInstruct~\cite{luo2025graphinstructempoweringlargelanguage}, 
NLGraph~\cite{wang2024languagemodelssolvegraph}, GPC~\cite{ICLR2025_9316da9c}, and 
LLM4DyG~\cite{zhang2024llm4dyglargelanguagemodels}.

\textbf{Tasks and Metrics}. We evaluate AdaSTORM across four core dynamic graph tasks:

Propagation (Reachability): Evaluates whether a node $v$ is temporally reachable from $u$ via a time-respecting path $\mathcal{P}_{u \to v} = \{(u_i, u_{i+1}, s_i, t_i)\}_{i=1}^k$, 
where $u_1 = u$, $u_{k+1} = v$, and $t_i \le s_{i+1}$.

Statistical (Temporal Motif Counting): Counts the total occurrences 
of a target temporal motif configuration $M_t$, defined as the cardinality 
$|\{\mathcal{G}' \subseteq \mathcal{G} \mid \mathcal{G}' \cong M_t\}|$.

Topological Analysis (Connected Components): Computes the number of connected components $|\mathcal{C}_c(\mathcal{G}_{\Delta T})|$ within a specified temporal 
window $\Delta T = [t_{\text{start}}, t_{\text{end}}]$.

Structural Mining (Community Detection): Discovers the latent community 
structures $|\mathcal{C}_d(\mathcal{G}_{\Delta T})|$ within the temporal window $\Delta T$.

We employ accuracy as the primary evaluation metric. For community detection, the output of the Label Propagation algorithm~\cite{PhysRevE.76.036106} serves 
as the ground truth.

\begin{table*}[t]
\centering
\caption{Performance comparison under various graph scales ($N$). We compare AdaSTORM with leading LLMs, chain- and debate-based multi-agent baselines. ``Acc'' denotes accuracy, ``$k$'' denotes the average number 
of regions generated by AdaSTORM's adaptive partitioning stage, and ``-'' indicates failure due to exceeding context window.}
\scriptsize
\setlength{\tabcolsep}{2.2pt}
\renewcommand{\arraystretch}{1.08}

\setlength{\aboverulesep}{0pt}
\setlength{\belowrulesep}{0pt}
\setlength{\cmidrulesep}{0pt}

\resizebox{\textwidth}{!}{
\begin{tabular}{lcccccccccccccccccccccccc}
\toprule
\rowcolor[HTML]{E6E6E6}
\cellcolor[HTML]{E6E6E6}
& \multicolumn{6}{c}{Community Detection}
& \multicolumn{6}{c}{Connected Components}
& \multicolumn{6}{c}{Reachability}
& \multicolumn{6}{c}{Temporal Motif Counting} \\[-0.5pt]

\arrayrulecolor[HTML]{E6E6E6}\cline{1-1}
\arrayrulecolor{black}
\cmidrule(l{0pt}r{0pt}){2-7}
\cmidrule(l{0pt}r{0pt}){8-13}
\cmidrule(l{0pt}r{0pt}){14-19}
\cmidrule(l{0pt}r{0pt}){20-25}

\rowcolor[HTML]{E6E6E6}
\multicolumn{1}{c}{\cellcolor[HTML]{E6E6E6}Models}
& \multicolumn{2}{c}{500}
& \multicolumn{2}{c}{800}
& \multicolumn{2}{c}{1000}
& \multicolumn{2}{c}{500}
& \multicolumn{2}{c}{800}
& \multicolumn{2}{c}{1000}
& \multicolumn{2}{c}{500}
& \multicolumn{2}{c}{800}
& \multicolumn{2}{c}{1000}
& \multicolumn{2}{c}{500}
& \multicolumn{2}{c}{800}
& \multicolumn{2}{c}{1000} \\[-0.5pt]

\arrayrulecolor[HTML]{E6E6E6}\cline{1-1}
\arrayrulecolor{black}
\cmidrule(l{0pt}r{0pt}){2-3}
\cmidrule(l{0pt}r{0pt}){4-5}
\cmidrule(l{0pt}r{0pt}){6-7}
\cmidrule(l{0pt}r{0pt}){8-9}
\cmidrule(l{0pt}r{0pt}){10-11}
\cmidrule(l{0pt}r{0pt}){12-13}
\cmidrule(l{0pt}r{0pt}){14-15}
\cmidrule(l{0pt}r{0pt}){16-17}
\cmidrule(l{0pt}r{0pt}){18-19}
\cmidrule(l{0pt}r{0pt}){20-21}
\cmidrule(l{0pt}r{0pt}){22-23}
\cmidrule(l{0pt}r{0pt}){24-25}

\rowcolor[HTML]{E6E6E6}
\cellcolor[HTML]{E6E6E6}
& Acc. & $k$
& Acc. & $k$
& Acc. & $k$
& Acc. & $k$
& Acc. & $k$
& Acc. & $k$
& Acc. & $k$
& Acc. & $k$
& Acc. & $k$
& Acc. & $k$
& Acc. & $k$
& Acc. & $k$ \\
\midrule

DeepSeek-Distill-7B
& 0\% & 1.0
& -- & --
& -- & --
& 0\% & 1.0
& -- & --
& -- & --
& 47\% & 1.0
& -- & --
& -- & --
& 0\% & 1.0
& -- & --
& -- & -- \\

\rowcolor[HTML]{EDEDED}
DeepSeek-Distill-14B
& 0\% & 1.0
& -- & --
& -- & --
& 0\% & 1.0
& -- & --
& -- & --
& 63\% & 1.0
& -- & --
& -- & --
& 0\% & 1.0
& -- & --
& -- & -- \\

DeepSeek-Distill-32B
& 0\% & 1.0
& -- & --
& -- & --
& 0\% & 1.0
& -- & --
& -- & --
& 66\% & 1.0
& -- & --
& -- & --
& 0\% & 1.0
& -- & --
& -- & -- \\

\rowcolor[HTML]{EDEDED}
GPT-4o mini
& 0\% & 1.0
& 0\% & 1.0
& 0\% & 1.0
& 0\% & 1.0
& 0\% & 1.0
& 0\% & 1.0
& 52\% & 1.0
& 48\% & 1.0
& 46\% & 1.0
& 0\% & 1.0
& 0\% & 1.0
& 0\% & 1.0 \\

DeepSeek-V4-Flash
& 0\% & 1.0
& 0\% & 1.0
& 0\% & 1.0
& 0\% & 1.0
& 0\% & 1.0
& 0\% & 1.0
& 58\% & 1.0
& 65\% & 1.0
& 53\% & 1.0
& 0\% & 1.0
& 0\% & 1.0
& 0\% & 1.0 \\

\rowcolor[HTML]{EDEDED}
Chain-GPT-4o mini
& 0\% & 1.0
& 0\% & 1.0
& 0\% & 1.0
& 0\% & 1.0
& 0\% & 1.0
& 0\% & 1.0
& 66\% & 1.0
& 68\% & 1.0
& 63\% & 1.0
& 0\% & 1.0
& 0\% & 1.0
& 0\% & 1.0 \\

Debate-GPT-4o mini
& 0\% & 1.0
& 0\% & 1.0
& 0\% & 1.0
& 0\% & 1.0
& 0\% & 1.0
& 0\% & 1.0
& 62\% & 1.0
& 56\% & 1.0
& 53\% & 1.0
& 0\% & 1.0
& 0\% & 1.0
& 0\% & 1.0 \\

\midrule
\rowcolor[HTML]{EDEDED}
Ours-DeepSeek-Distill-7B
& 23\% & 20.9
& 16\% & 32.4
& 10\% & 46.1
& 58\% & 48.0
& \underline{56\%} & 79.4
& 42\% & 106.6
& 57\% & 14.6
& 51\% & 20.5
& 52\% & 27.5
& \textbf{100\%} & 5.1
& 82\% & 9.7
& 56\% & 18.8 \\

Ours-DeepSeek-Distill-14B
& \underline{45\%} & 16.1
& \underline{30\%} & 33.3
& \underline{19\%} & 47.4
& \underline{62\%} & 20.4
& 43\% & 35.9
& \underline{45\%} & 41.9
& \underline{76\%} & 12.0
& \underline{70\%} & 18.7
& \underline{66\%} & 25.4
& \textbf{100\%} & 5.0
& \underline{86\%} & 9.1
& \underline{60\%} & 17.3 \\

\rowcolor[HTML]{D7F6FF}
Ours-DeepSeek-Distill-32B
& \textbf{71\%} & 7.5
& \textbf{52\%} & 95.4
& \textbf{26\%} & 162.0
& \textbf{90\%} & 18.0
& \textbf{77\%} & 27.2
& \textbf{63\%} & 34.9
& \textbf{90\%} & 9.4
& \textbf{88\%} & 15.1
& \textbf{72\%} & 22.5
& \textbf{100\%} & 5.0
& \textbf{91\%} & 7.0
& \textbf{83\%} & 12.1 \\

\bottomrule
\end{tabular}
}
\label{table:main}
\end{table*}
\textbf{Datasets}. Synthetic dynamic graphs are generated by first initializing 
a static topology via the Stochastic Block Model (SBM)~\cite{abbe2023communitydetectionstochasticblock}, $G_0 = \text{SBM}(N, C, p_{\text{in}}, p_{\text{out}})$. Each edge is assigned a random timestamp uniformly distributed across the total time span, defined as $\alpha \times |E_0|$ (defaulting to $\alpha = 1.5$). To simulate edge dynamics, all initial edges are designated as add operations , with delete operations subsequently introduced to a random subset, yielding the dynamic graph quadruple representation $(u, v, s, t)$. For reachability tasks, datasets are generated to ensure a balanced distribution of positive and negative samples. Full reproducibility is secured via a fixed random seed. 

To evaluate generalization, we further employ five real-world dynamic graph datasets with richer structural and temporal diversity than synthetic benchmarks:  \textit{Wikipedia}~\cite{Kumar_2019}, an interaction network with 9,227 nodes 
(editors and Wikipedia pages) and 157,474 timestamped edit edges with 172-dimensional 
features; \textit{Reddit}~\cite{poursafaei2022betterevaluationdynamiclink}, a social 
network consisting of 10,984 nodes (users and subreddits) and 672,447 timestamped 
posting edges with 172-dimensional features; \textit{Enron}~\cite{Shetty2004TheEE}, 
a communication network with 184 nodes (employees) and 125,235 timestamped email 
edges; \textit{Flights}~\cite{6846743}, an airport network consisting of 13,169 nodes 
(airports) and 1,927,145 timestamped flight edges; and \textit{UNTrade}~\cite{MacDonald2015RethinkingAT}, a trade network consisting of 255 nodes (countries) and 507,497 timestamped trade edges. We sample dense local subgraphs with random walk with restart (RWR)~\cite{6482622}, selecting the top-$k$ most frequently visited nodes from random sources to preserve local topology.

\textbf{Parameter Settings}. All evaluations are executed via official DeepSeek APIs~\cite{Guo_2025} 
under default configurations. The warm-start graph partitioner defaults to METIS~\cite{781339} with $k_0 = 5$. Detailed parameter configurations are provided in Appendix~\ref{app:parameters}.

\subsection{Performance on Large-Scale Dynamic Graphs (RQ1)}
To evaluate AdaSTORM under substantial structural complexity, we generate 100 
synthetic dynamic graphs for each scale ($N \in \{500, 800, $ $1000\}$). All evaluations are repeated five times, with the average metrics reported in 
Table~\ref{table:main}. Notably, for Reachability task, AdaSTORM provides correct temporal paths to justify 
its answers, while baselines likely rely on guesswork. We summarize the core empirical findings below:

\textbf{Observation 1: Breaking through the Scaling Bottleneck}. Facing massive 
dynamic graph topologies, even leading closed-source models and multi-agent baselines 
suffer catastrophic failure, yielding near-zero accuracy except for Reachability 
tasks. This demonstrates that such complexity \textbf{far exceeds} the reasoning 
capabilities of a standalone agent. Furthermore, standard multi-agent paradigms 
(e.g., chain- or debate-based) 
fail to handle the combinatorial explosion triggered by graph scale~\cite{Lu2025KARMALM,fourney2024magenticonegeneralistmultiagentsolving,xu2025languageagentsreinforcementlearning}, as they typically operate globally on the entire graph topology while \textbf{lacking explicit mechanisms} for large-scale graph partitioning.
In contrast, AdaSTORM effectively \textbf{breaks through} the scaling bottleneck, 
successfully scaling reasoning capabilities to thousand-node networks, while others 
can only handle graphs with fewer than a hundred nodes. Meanwhile, it significantly 
\textbf{outperforms} all five single-agent and two multi-agent baselines, notably 
achieving 100\% accuracy on the Temporal Motif Counting task at $N = 500$, marking 
a fundamental advancement in LLM-based graph reasoning.

\textbf{Observation 2: Consistent Performance Gains across Model Scales}. AdaSTORM delivers robust performance improvements across varying backbone scales, consistently outperforming leading closed-source models, effectively unlocking the latent 
capabilities of smaller models in graph reasoning tasks. This advantage holds firmly across all evaluated tasks and graph sizes, validating the structural robustness of AdaSTORM. Notably, the larger 32B backbone yields superior performance owing to its greater inherent capacity.

\textbf{Observation 3: Inherent Scalability via Scale-Agnostic Partitioning}. AdaSTORM demonstrates seamless scalability to larger graph topologies. Since the adaptive partitioner is scale-agnostic by design, the multi-agent orchestration naturally bypasses conventional context window constraints and scale bottlenecks, endowing the architecture with inherent scaling capabilities.

\subsection{Performance on Existing Benchmarks (RQ2)}
\begin{figure*}[t]
    \centering

    \begin{minipage}[b]{0.25\textwidth}
        \centering
        \includegraphics[width=\linewidth]{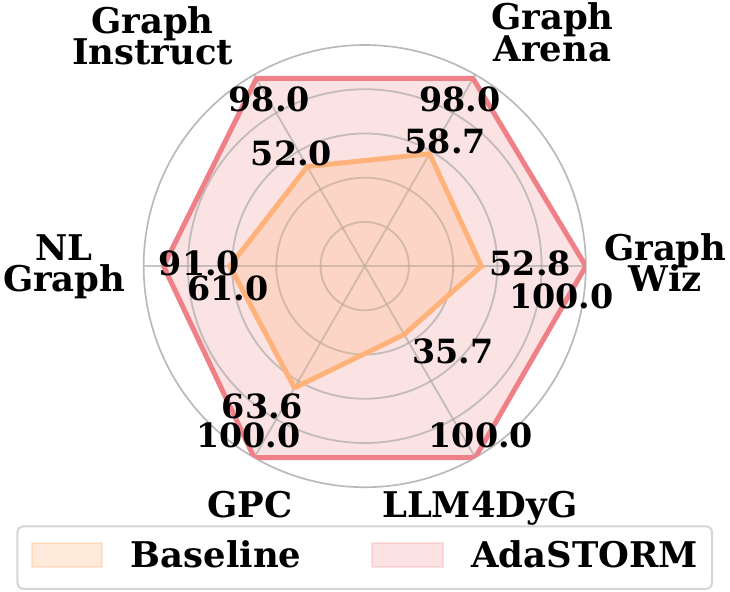}
        \captionof{figure}{Performance comparison on existing graph reasoning 
        benchmarks (Ours-DeepSeek-Distill-32B).
        }
        \label{fig:radar}
    \end{minipage}
    \hfill 
    %
    \begin{minipage}[b]{0.40\textwidth}
        \centering
        \includegraphics[width=\linewidth]{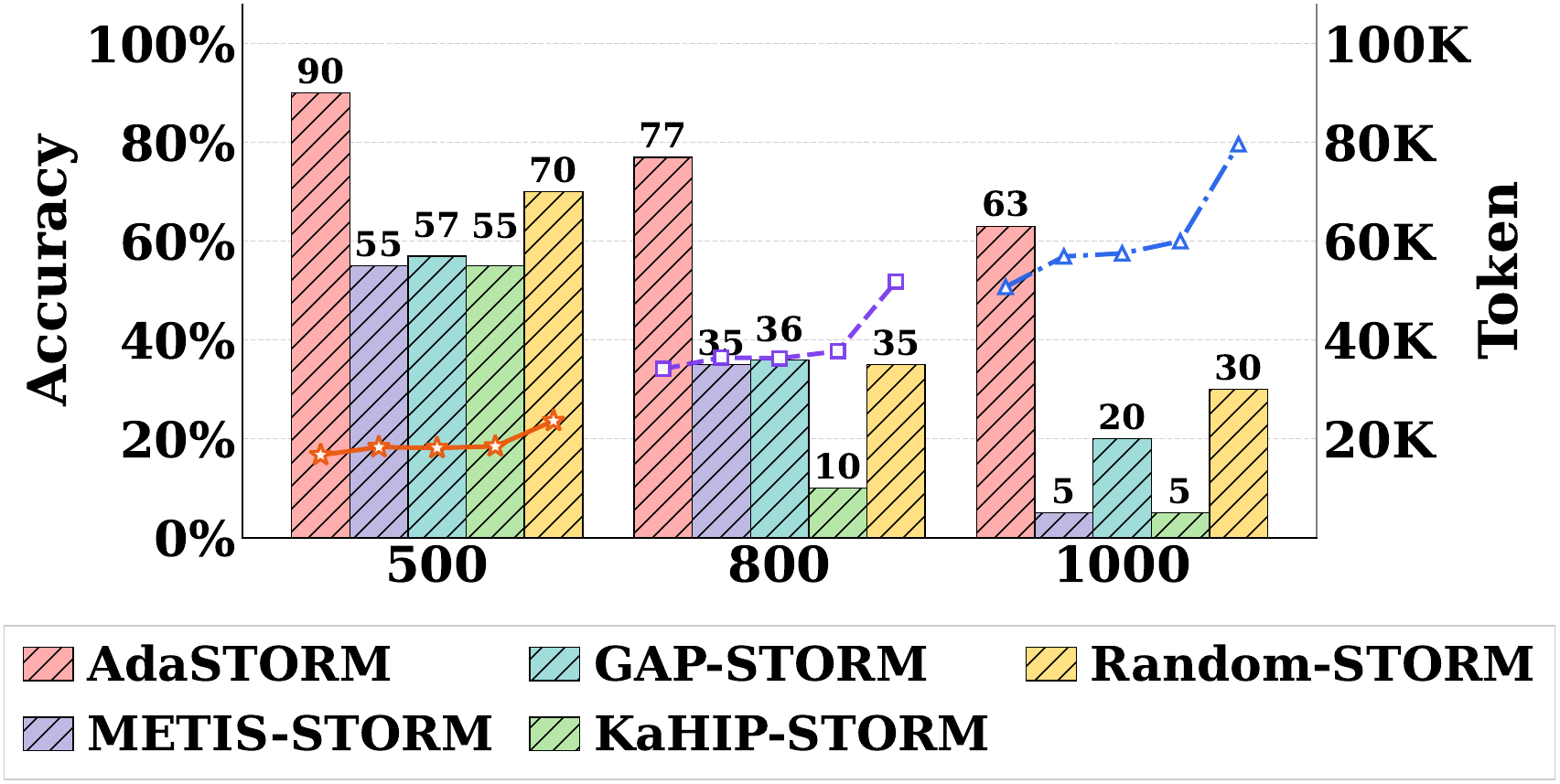}
        \captionof{figure}{Ablation results (performance and token cost) 
        of the adaptive partitioner across graph scales. Evaluated 
        on the connected components task under same partition counts.}
        \label{fig:ablation}
    \end{minipage}
    \hfill
    %
    \begin{minipage}[b]{0.28\textwidth}
        \centering
        \includegraphics[width=\linewidth]{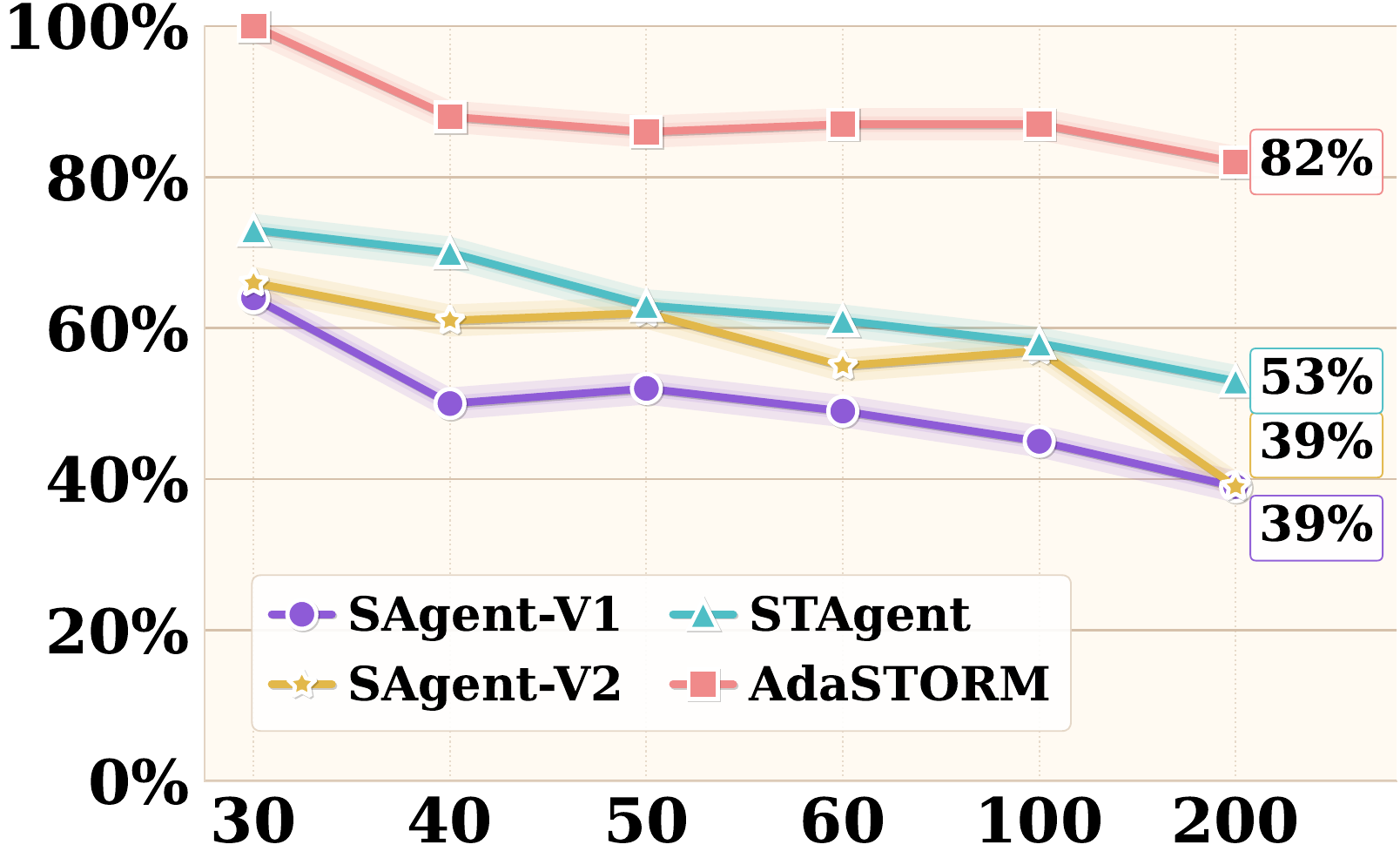}
        \captionof{figure}{Ablation results of quadruple representation, 
        spatio-temporal decoupling, and multi-agent orchestration on the 
        reachability task.}
        \label{fig:broken_line}
    \end{minipage}

\end{figure*}

AdaSTORM is evaluated on recent graph reasoning benchmarks under their native task 
and graph configurations. Performance metrics are compared against state-of-the-art (SOTA) results reported in their original papers, as plotted in Fig.~\ref{fig:radar}. The empirical findings show that AdaSTORM efficiently resolves a diverse array of small-to-medium scale graph tasks across all established benchmarks.

\begin{table}[t]
\centering
\caption{Dataset statistics and generalization evaluation on real-world networks.}
\label{tab:real_world_results}
\scriptsize
\setlength{\tabcolsep}{2.0pt}
\renewcommand{\arraystretch}{1.08}

\setlength{\aboverulesep}{0pt}
\setlength{\belowrulesep}{0pt}
\setlength{\cmidrulesep}{0pt}

\resizebox{\columnwidth}{!}{
\begin{tabular}{lcccccccccc}
\toprule
\rowcolor[HTML]{E6E6E6}
\multicolumn{1}{c}{\cellcolor[HTML]{E6E6E6}}
& \multicolumn{1}{c}{\cellcolor[HTML]{E6E6E6}}
& \multicolumn{1}{c}{\cellcolor[HTML]{E6E6E6}}
& \multicolumn{2}{c}{Community}
& \multicolumn{2}{c}{Connected}
& \multicolumn{2}{c}{Reachability}
& \multicolumn{2}{c}{Temporal Motif} \\[-0.2pt]

\rowcolor[HTML]{E6E6E6}
\multicolumn{1}{c}{\cellcolor[HTML]{E6E6E6}Dataset}
& \multicolumn{1}{c}{\cellcolor[HTML]{E6E6E6}\#Nodes}
& \multicolumn{1}{c}{\cellcolor[HTML]{E6E6E6}\#Edges}
& \multicolumn{2}{c}{Detection}
& \multicolumn{2}{c}{Components}
& \multicolumn{2}{c}{}
& \multicolumn{2}{c}{Counting} \\

\arrayrulecolor[HTML]{E6E6E6}\cline{1-3}
\arrayrulecolor{black}
\cmidrule(l{0pt}r{0pt}){4-5}
\cmidrule(l{0pt}r{0pt}){6-7}
\cmidrule(l{0pt}r{0pt}){8-9}
\cmidrule(l{0pt}r{0pt}){10-11}

\rowcolor[HTML]{E6E6E6}
\multicolumn{1}{c}{\cellcolor[HTML]{E6E6E6}}
& \multicolumn{1}{c}{\cellcolor[HTML]{E6E6E6}}
& \multicolumn{1}{c}{\cellcolor[HTML]{E6E6E6}}
& Acc. & $k$
& Acc. & $k$
& Acc. & $k$
& Acc. & $k$ \\
\midrule

Reddit  & 500 & 3,025   & 35\% & 32.0 & 73\% & 8.1  & 65\% & 8.8  & 100\% & 9.7  \\
\rowcolor[HTML]{EDEDED}
Enron   & 184 & 125,235 & 54\% & 79.0 & 95\% & 5.0  & 70\% & 28.0 & 10\%  & 28.0 \\
Flight  & 300 & 15,306  & 20\% & 8.0  & 91\% & 5.0  & 65\% & 30.6 & 15\%  & 5.2  \\
\rowcolor[HTML]{EDEDED}
UNtrade & 255 & 507,497 & 21\% & 8.0  & 93\% & 5.0  & 50\% & 45.4 & 5\%   & 5.0  \\
Wiki    & 500 & 2,136   & 15\% & 53.7 & 15\% & 11.4 & 76\% & 14.0 & 100\% & 14.1 \\
\bottomrule
\end{tabular}
}
\vspace{-3mm}
\end{table}
\begin{table}[t]
\centering
\caption{Ablation results of the capacity estimator under different graph scales. ``w/o CE'' denotes replacing the Capacity Estimator with a random feasibility score.}
\label{tab:ce_ablation}
\scriptsize
\setlength{\tabcolsep}{3.2pt}
\renewcommand{\arraystretch}{1.08}

\setlength{\aboverulesep}{0pt}
\setlength{\belowrulesep}{0pt}
\setlength{\cmidrulesep}{0pt}

\resizebox{\columnwidth}{!}{
\begin{tabular}{lccccccccc}
\toprule
\rowcolor[HTML]{E6E6E6}
\multicolumn{1}{c}{\cellcolor[HTML]{E6E6E6}}
& \multicolumn{3}{c}{500}
& \multicolumn{3}{c}{800}
& \multicolumn{3}{c}{1000} \\

\arrayrulecolor[HTML]{E6E6E6}\cline{1-1}
\arrayrulecolor{black}
\cmidrule(l{0pt}r{0pt}){2-4}
\cmidrule(l{0pt}r{0pt}){5-7}
\cmidrule(l{0pt}r{0pt}){8-10}

\rowcolor[HTML]{E6E6E6}
\multicolumn{1}{c}{\cellcolor[HTML]{E6E6E6}Method}
& Acc. & $k$ & Token
& Acc. & $k$ & Token
& Acc. & $k$ & Token \\
\midrule

AdaSTORM w/o CE
& 61\% & 34.0 & 33,756
& 54\% & 47.5 & 61,154
& 42\% & 51.2 & 114,580 \\
\rowcolor[HTML]{D7F6FF}
\textbf{AdaSTORM}
& \textbf{90\%} & \textbf{18.0} & \textbf{16,723}
& \textbf{77\%} & \textbf{27.2} & \textbf{34,146}
& \textbf{63\%} & \textbf{34.9} & \textbf{50,754} \\
\bottomrule
\end{tabular}
}

\end{table}
\subsection{Generalization on Real-World Datasets (RQ3)}
Evaluations are extended to real-world networks to test AdaSTORM against richer structural diversity than synthetic datasets. Specifically, we employ Ours-DeepSeek-Distill-32B as the backbone model to evaluate our framework across all datasets. The experimental results and dataset statistics are summarized in Table.~\ref{tab:real_world_results}. The outcomes demonstrate that AdaSTORM generalizes robustly to real-world environments, underscoring its practical viability.




\subsection{Ablation Study}

\textbf{Capacity Estimator}. To verify its efficacy, we substitute the Capacity 
Estimator with a random feasibility score while keeping the rest of the pipeline 
unchanged. Based on the connected components task and employing Ours-DeepSeek-Distill-32B as the backbone model, The results in Table~\ref{tab:ce_ablation} demonstrate that the Capacity Estimator 
provides effective guidance for AdaSTORM.

\textbf{Adaptive Partitioner}. To isolate its specific contribution, we replace the 
adaptive partitioner with alternative partitioning algorithms: METIS~\cite{781339}, GAP~\cite{nazi2019gapgeneralizableapproximategraph}, KaHIP~\cite{Akhremtsev2014KaHIPK}, and Random partitioning, while the subregion count to the value determined by our method. This yields four evaluation variants: METIS-STORM, GAP-STORM, KaHIP-STORM, and Random-STORM. Their performance metrics and token overhead are compared in Fig.~\ref{fig:ablation}. The results demonstrate that our adaptive partitioner not only automates region count determination but also yields superior topological quality, delivering higher end-to-end performance and lower token overhead under an identical partition count.

\textbf{Dynamic Graph Quadruple Representation $(u, v, s, t)$}. The dynamic graph is formalized within the prompt using the quadruple $(u, v, s, t)$. Compared to existing $(u, v, t, a/d)$~\cite{hao2026llmtmbenchmarkingoptimizingllms}, it maximizes information density with minimal token overhead, facilitating LLM 
comprehension. To isolate the specific contribution of this design, we evaluate two standalone agent variants on the Reachability task: SAgent-v1 employing $(u, v, t, a/d)$ and SAgent-v2 utilizing $(u, v, s, t)$. The experimental results, illustrated in Fig.~\ref{fig:broken_line}, validate the necessity of this representation optimization.

\textbf{Spatio-Temporal Decoupling}. Prior studies~\cite{zhang2024llm4dyglargelanguagemodels} have validated the efficacy of prompt-level spatio-temporal decoupling in single LLMs. To evaluate this mechanism under a multi-agent setting, we introduce a variant STAgent, which orchestrates decoupled spatial and temporal agents, and compare it against a monolithic single-agent baseline. As shown in Fig.~\ref{fig:broken_line}, STAgent consistently outperforms the single-agent baseline, fully validating the architectural superiority of explicit spatio-temporal decoupling.

\textbf{Multi-Agent Orchestration}. Incorporating the adaptive partitioning mechanism into the spatial agent of STAgent yields our complete framework. As shown in Fig.~\ref{fig:broken_line}, comparing AdaSTORM against the STAgent variant on the reachability task, validates the efficacy of multi-agent 
orchestration.

\section{Related Work}


\textbf{LLMs for Graphs}. LLM-based graph analytics generally bifurcates into benchmarking and methodological development. Benchmarks span static environments (e.g., GraphInstruct~\cite{luo2025graphinstructempoweringlargelanguage}, GraphArena~\cite{tang2025grapharenaevaluatingexploringlarge}, GraphOmni~\cite{xu2026graphomnicomprehensiveextensiblebenchmark}) and temporal dynamics (e.g., LLM4DyG~\cite{zhang2024llm4dyglargelanguagemodels}, LLMTM~\cite{hao2026llmtmbenchmarkingoptimizingllms}). Methodologically, existing efforts fall into two paradigms: (i) GNN-enhanced LLMs~\cite{tang2024graphgptgraphinstructiontuning,Zhang2024GraphTranslatorAG}, 
injecting structural embeddings into token spaces yet limiting cross-task generalization; 
and (ii) Text-flattened LLMs~\cite{ouyang2024gundamaligninglargelanguage,11222956,fatemi2023talklikegraphencoding,perozzi2024letgraphtalkingencoding}, 
serializing graphs into text descriptions for direct reasoning. While monolithic approaches scale poorly due to exponential reasoning overhead and finite context windows, AdaSTORM bypasses these architectural limits via adaptive multi-agent orchestration, scaling dynamic graph reasoning to unprecedented network dimensions.


\textbf{Multi-Agent Systems}. Multi-agent systems (MAS)~\cite{hong2024metagptmetaprogrammingmultiagent,wu2026talkrightspecialistsiterative} 
orchestrate specialized LLM-based agents to achieve autonomous problem-solving across 
diverse domains, spaning from software development to embodied control~\cite{qian2024experientialcolearningsoftwaredevelopingagents,guo2024embodiedllmagentslearn, chen2024scalablemultirobotcollaborationlarge, 10610855}. Technically, these frameworks orchestrate specialized agents through structured 
topologies, such as chains, trees, or graphs~\cite{qian2025scalinglargelanguagemodelbased}. This topological alignment renders MAS inherently compatible with dynamic graph reasoning.




\textbf{Graph Partitioning}. Conventional graph partitioning methodologies treat the problem as NP-hard discrete optimization, typically resolved via heuristics or approximations~\cite{748202, 9938438,10.1145/2339530.2339722}. However, these approaches rely on handcrafted objectives, suboptimally utilizing 
node attributes. While neural partitioning methods incorporate rich features and 
clustering constraints via continuous graph cuts~\cite{nazi2019gapgeneralizableapproximategraph,10.1016/j.patcog.2022.109126}, 
modularity maximization~\cite{2110eb02c20447e0b13da6f62c7ae963,tsitsulin2023graphclusteringgraphneural,bhowmick2023dgclusterneuralframeworkattributed}, 
or combinatorial reformulations~\cite{jung2022learningsolveminimumcost,shah2024neurocutneuralapproachrobust}, their reliance on static partition counts fails to accommodate dynamic multi-agent orchestration. To break this bottleneck, we introduce an adaptive partitioner that dynamically 
determining the optimal region count on-the-fly while preserving superior structural 
partition quality.


\section{Conclusion}
This work introduces AdaSTORM, a pioneering multi-agent framework tailored for large-scale dynamic graph reasoning. By coupling adaptive partitioning with collective reasoning, the architecture orchestrates 
multi-agent collaboration for dynamic graph reasoning. Empirical evaluations demonstrate that AdaSTORM successfully breaks through
the scaling bottlenecks, scaling reasoning to thousand-node networks and
significantly outperforms seven competitive baselines. Furthermore, it achieves state-of-the-art accuracy on existing benchmarks 
and generalizes robustly to real-world datasets, underscoring its practical viability.

As a modular framework, AdaSTORM inherently enhances the scalability of alternative graph-reasoning methods while adapting to diverse partitioning objectives. Future work will explore instruction tuning within this orchestration to unlock further performance gains.
\clearpage
\section{Limitations}

Although AdaSTORM demonstrates strong scalability and effectiveness on large-scale dynamic graph reasoning, its current formulation is most naturally suited to tasks where global graph reasoning can be decomposed into capacity-constrained subregions and recomposed through inter-region collaboration. This setting aligns well with the topological and temporal reasoning tasks considered in this work, but may be less directly applicable to scenarios whose solutions depend heavily on rich semantic attributes, external domain knowledge, or highly ambiguous natural-language objectives beyond the structural and temporal information encoded in the graph. In addition, AdaSTORM treats the backbone LLM as a black-box reasoner and improves its reasoning capability through adaptive partitioning and multi-agent orchestration rather than model-level adaptation. Therefore, the overall performance can still be affected by local reasoning errors, unstable intermediate outputs, or imperfect capacity estimation.

\section{Impact Statement}

This work aims to improve the ability of LLMs to reason over large-scale dynamic graphs, a setting that appears in many real-world domains such as social interaction analysis, communication networks, transportation systems, and temporal knowledge discovery. By decomposing complex graph structures into capacity-matched subregions and coordinating multiple agents for spatio-temporal reasoning, AdaSTORM provides a scalable framework for applying LLM-based reasoning to graph scenarios that are difficult to handle with standalone models. This may benefit research on graph analytics, multi-agent systems, and LLM-based decision support by enabling more accessible reasoning over structurally complex and temporally evolving data. At the same time, dynamic graphs may contain sensitive relational or behavioral information, and automated reasoning over such data should be conducted with appropriate privacy protection, data governance, and human oversight. The proposed framework is intended as a general reasoning methodology, and its deployment in high-stakes domains should be accompanied by careful validation and responsible use.


\bibliography{reference}
\clearpage
\appendix
\appendix

\section{Additional Preliminaries}
\label{app:preliminaries}

\textbf{Graph Partitioning}. Given a graph $G = (\mathcal{V}, E_s)$, a graph 
partitioning $\mathcal{P}$ is defined as a collection of $k$ mutually disjoint 
and localized subgraphs:
\begin{equation}
\mathcal{P} = \{P_1, P_2, \ldots, P_k\}.
\end{equation}
Each partition is formalized as a subgraph $P_i = (V_i, E_i)$. The vertex sets 
$\{V_1, V_2, \ldots, V_k\}$ form a valid partition over the global node set $\mathcal{V}$, 
satisfying:
\begin{equation}
V_i \cap V_j = \emptyset, \quad \forall i \neq j,
\qquad
\bigcup_{i=1}^{k} V_i = \mathcal{V}.
\end{equation}
The corresponding edge set $E_i$ is induced directly from the global edge set $E_s$, 
retaining all links strictly confined within $V_i$:
\begin{equation}
E_i = \bigl\{ \{u, v\} \in E_s \;\big|\; u \in V_i, \; v \in V_i \bigr\}.
\end{equation}

Unlike existing partitioning methods, the number of subregions $k$ in AdaSTORM 
is not pre-fixed, but adaptively determined according to the reasoning task 
and the target model's capacity.


\textbf{Cut Edges}. Given a graph $G = (\mathcal{V}, E_s)$ and a partitioning 
$\mathcal{P} = \{P_1, \ldots, P_k\}$, the pairwise cut-edge set between two 
distinct partitions $P_i$ and $P_j$ is defined as:
\begin{equation}
E_{\mathrm{cut}}(P_i, P_j) = \bigl\{ \{u, v\} \in E_s \;\big|\; u \in V_i, \; v \in V_j \bigr\},
\end{equation}
where each edge in the undirected graph is represented as an unordered pair 
$\{u, v\}$. By aggregating these crossings, the global cut-edge set of the partitioning $\mathcal{P}$ is formulated as:
\begin{equation}
E_{\mathrm{cut}}(\mathcal{P}) = \bigcup_{1 \leq i < j \leq k} E_{\mathrm{cut}}(P_i, P_j).
\end{equation}

\textbf{Dynamic Mapping}. When temporal information is required, the corresponding 
dynamic event set $\mathcal{E}_{\mathrm{rec}}(E')$ is defined by recovering all 
lifecycle quadruplets from the global sequence $\mathcal{E}$:
\begin{equation}
\mathcal{E}_{\mathrm{rec}}(E') = \bigl\{ (u, v, s, t) \in \mathcal{E} \;\big|\; \{u, v\} \in E' \bigr\}.
\end{equation}

\section{Model Profiles}
\label{app:model_profiles}

To construct the explicit semantic attributes required by the Cognitive Estimator, 
we employ an auxiliary LLM to synthesize structured model profiles. 
Specifically, the auxiliary model is prompted to extracts inherent architectural histories, capacities, and empirical graph-task performance directly from runtime execution logs, thereby 
characterizing high-level semantic attributes, including model capabilities and operational scenarios, as detailed below:

\begin{itemize}
    \item \textbf{DeepSeek-R1-Distill-Qwen-7B}: This variant is an official DeepSeek-R1 
    distilled checkpoint built upon the Qwen2.5-Math-7B backbone. The R1 distilled 
    family is fine-tuned on an 800k-sample dataset curated directly from DeepSeek-R1. 
    Empirical evaluations indicate that it demonstrates robust proficiency in local reasoning, achieving approximately 79\% accuracy on Reachability and 71\% on Temporal Motif Counting, whereas it exhibits prominent bottlenecks on global topological tasks, dropping to 55\% on Connected Components and 43\% on Community Detection.
    
    \item \textbf{DeepSeek-R1-Distill-Qwen-14B}: This checkpoint is an official 
    DeepSeek-R1 distilled model grounded on the Qwen2.5-14B architecture, leveraging 
    the identical R1 distillation pipeline and the 800k-sample training configuration. 
    According to the execution logs, it delivers a more balanced topological reasoning 
    profile across the evaluation suite, elevating performance to approximately 82\% 
    on Reachability, 82\% on Temporal Motif Counting, 76\% on Connected Components, and 55\% on Community Detection.
    
    \item \textbf{DeepSeek-R1-Distill-Qwen-32B}: This architecture represents an official 
    DeepSeek-R1 distilled checkpoint based on the Qwen2.5-32B backbone. Official release 
    notes highlight that this 32B distilled variant performs on par with or surpasses 
    OpenAI o1-mini on selected reasoning benchmarks. Empirical logs demonstrate that it 
    establishes the upper accuracy frontier among the three models, securing approximately 
    89\% on Reachability, 90\% on Temporal Motif Counting, 80\% on Connected Components, and 
    71\% on Community Detection.
\end{itemize}

\section{Extended Task-Specific Performance Landscapes}
\label{app:extended_profiles}
\begin{figure}[!htbp]
    \centering
    \begin{subfigure}{0.43\columnwidth}
        \centering
        \includegraphics[width=\linewidth]{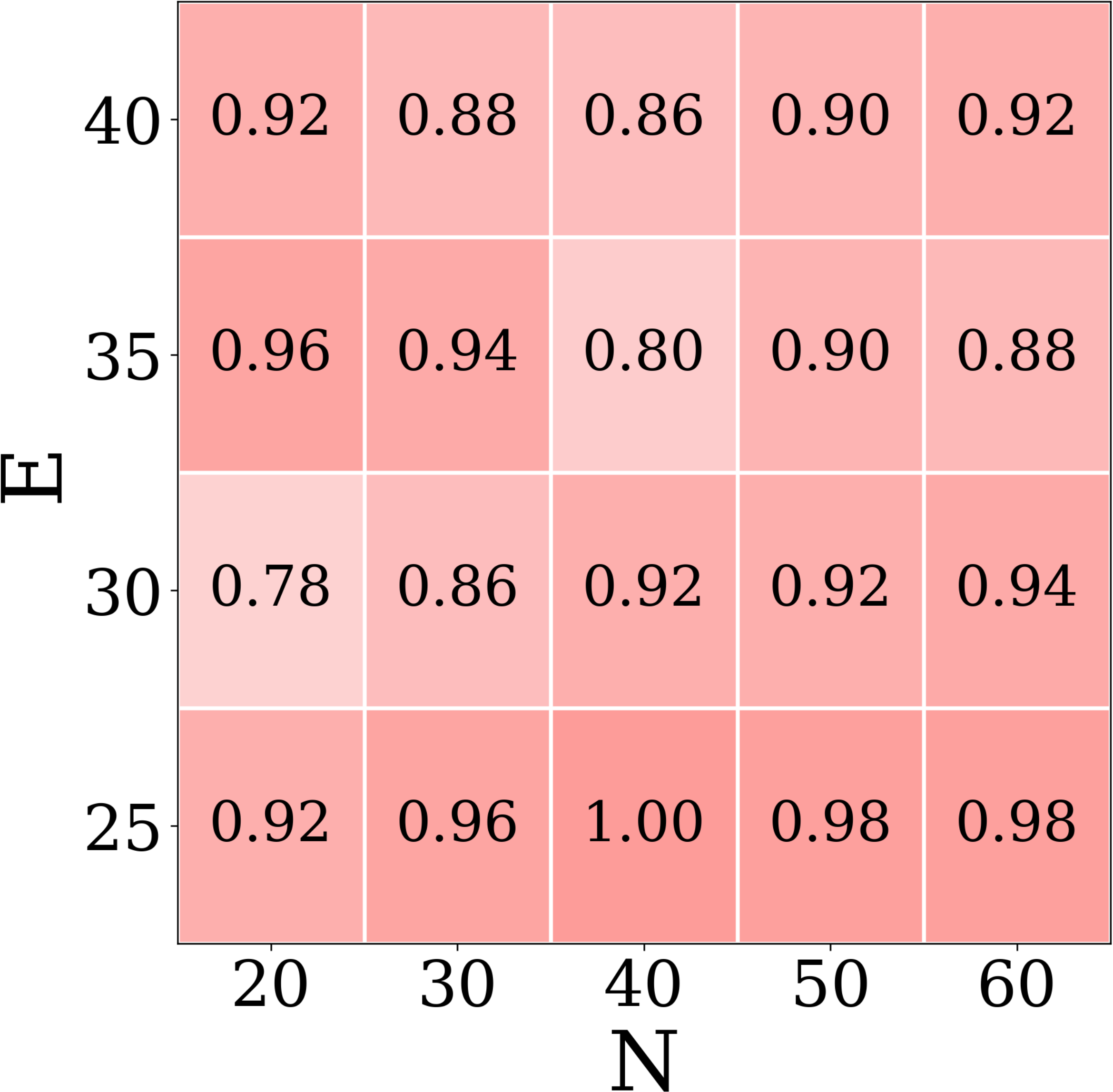}
        \caption{Graph reachability.}
        \label{fig:app_reachability_heatmap}
    \end{subfigure}
    \hfill 
    \begin{subfigure}{0.50\columnwidth}
        \centering
        \includegraphics[width=\linewidth]{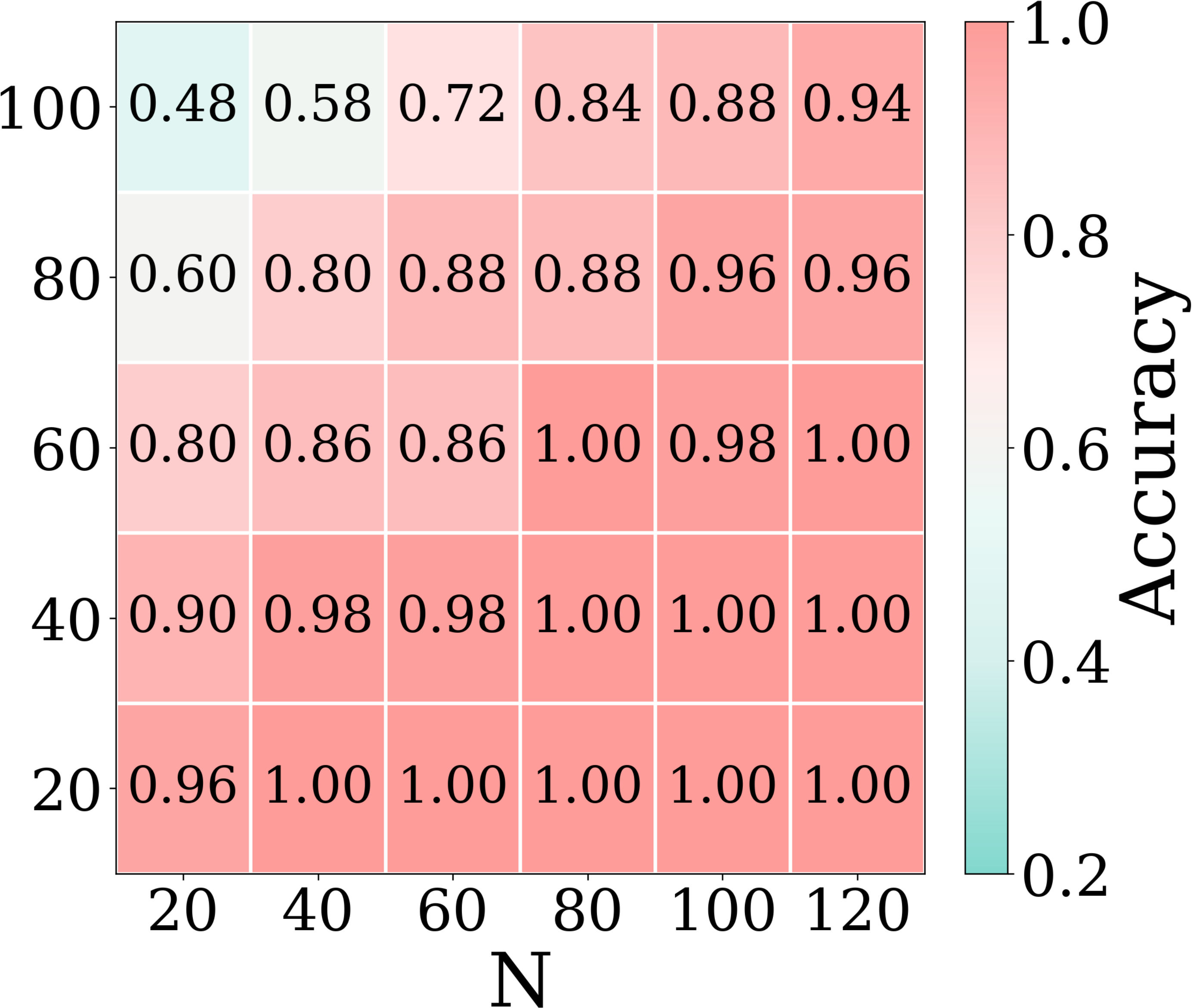}
        \caption{Temporal motif.}
        \label{fig:app_motif_heatmap}
    \end{subfigure}
    \caption{Reasoning performance of the identical target model under varying node ($N$) and edge ($E$) scales for extended tasks.}
    \label{fig:app_extended_heatmaps}
\end{figure}

Figure~\ref{fig:app_reachability_heatmap} 
and Figure~\ref{fig:app_motif_heatmap} provide extended performance evaluations for graph 
reachability and temporal motif detection. Under the identical model configuration, the 
resulting accuracy landscapes exhibit fundamentally distinct degradation trajectories as 
topological scales change. These evaluations directly corroborate our main claim that 
reasoning performance varies drastically across diverse tasks, rendering a universal, static partition count $k$ sub-optimal.

\section{Detailed Parameter Configurations}
\label{app:parameters}

The GNN backbone is configured as a 2-layer GraphSAGE~\cite{hamilton2018inductiverepresentationlearninglarge} 
with a hidden dimension of 32. Positional embeddings employ a default of 35 anchor nodes. In Eq.~\eqref{eq:step_reward}, the weight $w_c$ adapts dynamically, scaling from 0.05 when any sub-region is infeasible to 2.0 when all sub-regions 
achieve feasibility; the remaining coefficients are set to $\lambda_\Phi = 100$ (for numerical scale alignment), $\lambda_{\text{split}} = 0.05$, and 
$\lambda_{\text{reject}} = 0.03$. The feasibility threshold for Eq.~\eqref{eq:global_infeasibility_gap} is set to $\tau = 0.63$. 
For Eq.~\eqref{eq:steady_state}, the coefficient is $c = 0.85$. For Eq.~\eqref{eq:final_reward}, 
the terminal parameters are configured as $\lambda_{\text{final}} = 150$, 
$B_{\text{final}} = 150$, and $\lambda_{\text{call}}^{\text{final}} = 0.1$. 
The policy network $\pi_\theta$ is optimized using Adam with a learning 
rate of 0.0001, a discount factor of $\gamma = 0.99$, and a trajectory length of $T = 2$.

\section{Details of the Adaptive Partitioner}
\label{app:adaptive_partitioner}
\subsection{MDP Formulation}

We formulate iterative partition refinement as a Markov Decision Process (MDP) defined by $(S,A,\rho,R,\gamma)$. 
The objective is to learn a policy that produces a partitioning $\mathcal{P}$ satisfying the cognitive feasibility threshold across all regions while minimizing communication overhead. 
Here, $S$ is the state space, $A$ is the hybrid action space covering node migration and region splitting, $\rho:S\times A\times S\rightarrow[0,1]$ denotes the state transition probability, $R:S\times A\rightarrow\mathbb{R}$ is the reward function, and $\gamma\in(0,1)$ is the discount factor.

\subsection{Initialization and Positional Embeddings}

Rather than starting from empty partitions, we initialize the process with a warm start. 
Specifically, we employ METIS~\cite{10.1145/2339530.2339722} to decompose the graph into $k_0$ initial clusters based on raw node features and positional embeddings. 
This pre-partitioning provides a structured starting point for subsequent optimization.

To preserve structural proximity, we adopt Lipschitz Embedding~\cite{Bourgain1985} and compute positional features through Random Walk with Restart (RWR)~\cite{10.1145/1052934.1052938}. 
Let $\mathcal{V}_{anc}=\{v_1,\ldots,v_\alpha\}\subseteq\mathcal{V}$ be a set of $\alpha$ anchor nodes. 
For each anchor $v_i$, the steady-state probability vector $\vec{r}_i$ is defined as
\begin{equation}
\vec{r}_i
=
c\tilde{\mathbf{W}}\vec{r}_i
+
(1-c)\vec{e}_i,
\label{eq:steady_state}
\end{equation}
where $\tilde{\mathbf{W}}$ is the transition matrix, $\vec{e}_i$ is the one-hot indicator for anchor $v_i$, and $c\in(0,1)$ controls the balance between walking to neighbors and restarting at the anchor. 
Each node $u$ is represented by
\begin{equation}
\mathrm{pos}(u)
=
[r_{1u},\ldots,r_{\alpha u}],
\end{equation}
where $r_{iu}$ is the $u$-th element of $\vec{r}_i$. 
We concatenate raw node features with positional encodings to obtain the initial embedding:
\begin{equation}
\mathrm{emb}_{init}(u)
=
\mathbf{X}[u]\parallel \mathrm{pos}(u),
\label{eq:initial_embedding}
\end{equation}
where $\parallel$ denotes concatenation.

\subsection{Graph Encoders}

Building upon Eq.~\eqref{eq:initial_embedding}, we employ Graph Neural Networks (GNNs)~\cite{hamilton2018inductiverepresentationlearninglarge} to capture topology-aware node representations for partition refinement. 
The node migration policy and the region splitting policy use the same message-passing architecture but maintain separate trainable parameters. 
For each branch $b\in\{\mathrm{mig},\mathrm{split}\}$, the input representation of node $u$ is initialized as
\begin{equation}
\mathbf{h}_{u,b}^{0}
=
\mathrm{emb}_{init}(u).
\end{equation}
At layer $l+1$, the node representation is updated by
\begin{equation}
\mathbf{h}_{u,b}^{l+1}
=
\mathbf{W}_{1,b}^{l}\mathbf{h}_{u,b}^{l}
+
\mathbf{W}_{2,b}^{l}
\cdot
\frac{1}{|\mathcal{N}_u|}
\sum_{u'\in\mathcal{N}_u}
\mathbf{h}_{u',b}^{l},
\label{eq:gnn}
\end{equation}
where $\mathcal{N}_u$ denotes the neighborhood of $u$, and $\mathbf{W}_{1,b}^{l},\mathbf{W}_{2,b}^{l}$ are trainable parameters.

\subsection{State and High-level Action Selection}

At step $t$, the partition status is denoted as $\mathcal{P}^t=\{P_1^t,\ldots,P_k^t\}$. 
The state is defined as
\begin{equation}
S^t
=
\{
S_1^t,\ldots,S_k^t,
\Phi(\mathcal{P}^t),
C_t
\},
\end{equation}
where
\begin{equation}
S_i^t
=
\{
\mathbf{h}_{v}^{L}
\mid
v\in P_i^t
\},
\label{eq:state}
\end{equation}
$\Phi(\mathcal{P}^t)$ is the global infeasibility gap, and $C_t$ is the current inter-region communication cost.

The action space $\mathcal{A}$ contains two refinement types: node migration and region splitting. We define the policy input as
\begin{equation}
\mathbf{q}^t =
\left[
\min_{i\in\mathcal{P}^t}\phi_i^t,\ 
\Phi(\mathcal{P}^t)
\right].
\end{equation}

A lightweight policy head first selects the refinement operation 
$o^t \in \mathcal{O}^t$, where $\mathcal{O}^t$ contains node migration and region splitting:
\begin{equation}
\begin{aligned}
\pi_{\theta}^{\mathrm{type}}(o^t \mid S^t)
&= \\
&\mathrm{Softmax}_{o^t\in\mathcal{O}^t}
\left(
\mathrm{MLP}_{\theta_a}(\mathbf{q}^t)
\right).
\end{aligned}
\end{equation}
To reduce the combinatorial explosion of the action space, we employ heuristic candidate selection to identify candidate nodes and bottleneck regions, while the RL policy focuses on high-level decision optimization.

\subsection{Node Migration}

When the node migration action is selected, we identify a structurally misplaced node and reassign it using the migration policy. 
Let $P(v)$ be the current partition assignment of node $v$, and let
\begin{equation}
\mathcal{N}_p(v)
=
\{u\in\mathcal{N}(v)\mid P(u)=p\}
\end{equation}
be the neighbors of $v$ in partition $p$. 
The structural misalignment of $v$ is quantified by
\begin{equation}
\mathrm{NodeScore}^t(v)
=
\frac{
\max_{p\neq P(v)}|\mathcal{N}_p(v)|
}{
|\mathcal{N}_{P(v)}(v)|
}
\cdot
\frac{1}{d_v+\epsilon},
\end{equation}
where $d_v$ is the node degree and $\epsilon$ is a stability constant. 
This score highlights fragile boundary nodes that dominate the total edge cut.

Conditioned on the selected node $v$, the compatibility between $v$ and each candidate partition $p\in\mathcal{P}^t$ is computed by aggregating interactions with neighbors in $p$:
\begin{equation}
\begin{aligned}
&\mathrm{PartScore}_{\theta}(p,v) = \\[-0.5ex]
&\hspace{1.5em}
\mathrm{AGG}_{u\in\mathcal{N}_p(v)}
\Big(
\mathrm{MLP}_{\theta_m}
\big(
\mathbf{h}_{v,\mathrm{mig}}^L
\parallel
\mathbf{h}_{u,\mathrm{mig}}^L
\big)
\Big).
\end{aligned}
\end{equation}
The migration action is sampled from
\begin{equation}
\begin{aligned}
&\pi_{\theta}^{\mathrm{mig}}(a^t=p\mid S^t,v) = \\[-0.5ex]
&\hspace{4.0em}
\frac{
\exp(\mathrm{PartScore}_{\theta}(p,v))
}{
\sum_{p'\in\mathcal{P}_{\mathrm{act}}^t}
\exp(\mathrm{PartScore}_{\theta}(p',v))
}.
\end{aligned}
\end{equation}

If node $v$ is migrated from partition $i$ to partition $j$, the transition is
\begin{equation}
P_i^{t+1}
\leftarrow
P_i^t\setminus\{v\},
\quad
P_j^{t+1}
\leftarrow
P_j^t\cup\{v\}.
\end{equation}

\subsection{Region Splitting}

When the splitting action is selected, we identify the region that contributes most to the global infeasibility gap. 
For each partition $P_i\in\mathcal{P}$, its contribution is defined as
\begin{equation}
c_i
=
|P_i|\cdot\max(0,\tau-\phi_i).
\end{equation}
The target region is selected as
\begin{equation}
r
=
\arg\max_{P_i\in\mathcal{P}}c_i.
\end{equation}

Given the target region $r$, the splitting policy uses the topology-aware representations from the splitting encoder(Eq.~\eqref{eq:gnn}). 
For each node $u\in r$, we construct
\begin{equation}
\mathbf{z}_u
=
\mathrm{MLP}
\left(
\mathbf{h}_{u,\mathrm{split}}^{L}
\parallel
\mathrm{AGG}_{v\in r}
(\mathbf{h}_{v,\mathrm{split}}^{L})
\parallel
\mathbf{s}_r
\right),
\end{equation}
where $\mathbf{s}_r$ encodes region-specific structural attributes, including partition size, internal node degree, boundary degree, boundary ratio, and the estimated feasibility rate $\phi_r$. 
The binary assignment policy is
\begin{equation}
\pi_{\theta}^{\mathrm{split}}(b_u^t=1\mid S^t,r,u)
=
\frac{
\exp(g_{u,1})
}{
\exp(g_{u,0})+\exp(g_{u,1})
},
\end{equation}

\begin{equation}
\mathbf{g}_u
=
\mathrm{MLP}_{\theta_s}(\mathbf{z}_u).
\end{equation}
Node-wise assignments execute a bipartition of $r$. 
The partition set is updated as
\begin{equation}
\mathcal{P}^{t+1}
\leftarrow
(\mathcal{P}^{t}\setminus\{r\})
\cup
\{r_1,r_2\},
\end{equation}
where
\begin{equation}
r_1=\{u\in r\mid \hat{y}_u=0\},
\quad
r_2=\{u\in r\mid \hat{y}_u=1\}.
\end{equation}
This operation increases the number of partitions by one and refines the global partitioning granularity.

\subsection{Reward Design}

The immediate reward is defined as
\begin{equation}
\begin{split}
r_t = \, 
&w_c(C_{t-1}-C_t)
+
\lambda_\Phi(\Phi_{t-1}-\Phi_t) \\
&-
\lambda_{\mathrm{split}}\mathbbm{1}_{\mathrm{split}}
-
\lambda_{\mathrm{reject}}\mathbbm{1}_{\mathrm{reject}},
\end{split}
\end{equation}
where $\mathbbm{1}_{\mathrm{split}}$ penalizes unnecessary split actions to prevent over-segmentation, and $\mathbbm{1}_{\mathrm{reject}}$ penalizes ineffective maneuvers, including splitting attempts with no feasibility gain and node migrations that degrade feasibility or fail to reduce communication cost.

To account for long-term partitioning quality, we use discounted returns:
\begin{equation}
D_t
=
\sum_{j=0}^{T-t}
\gamma^j r_{t+j},
\label{eq:reward_dt}
\end{equation}
where $T$ is the horizon length and $\gamma\in(0,1]$ is the discount factor.

At the end of an episode, we add a terminal reward:
\begin{equation}
\begin{split}
r_T^{\mathrm{final}}
=
&-\lambda_{\mathrm{final}}\Phi_T
+
B_{\mathrm{final}}\mathbbm{1}_{|\mathcal{I}_T|=0} \\
&-
\lambda_{\mathrm{call}}^{\mathrm{final}}
\max(0,C_T-C_0),
\end{split}
\label{eq:final_reward}
\end{equation}
where $\mathcal{I}_T=\{P_i\in\mathcal{P}\mid \phi_i<\tau\}$ denotes the set of infeasible partitions. 
The first term penalizes residual infeasibility, the second term rewards global feasibility, and the third term discourages communication degradation.

\subsection{Policy Optimization and Inference}

We optimize the policy network by maximizing the expected return. 
The policy parameters $\theta$ are updated via the REINFORCE gradient estimator~\cite{712192}. 
For an episode with horizon $T$, the policy objective
\begin{equation}
J(\pi_\theta)
=
\mathbb{E}_{\pi_\theta}
\left[
\sum_{t=0}^{T}
D_t
\right]
\end{equation}
is optimized through
\begin{equation}
\nabla_\theta J(\pi_\theta)
=
\mathbb{E}_{\pi_\theta}
\left[
\sum_{t=0}^{T}
D_t
\nabla_\theta
\log\pi_\theta(a_t\mid s_t)
\right].
\end{equation}
Here, $a_t$ includes both node migration and region splitting. 
For splitting actions, $\log\pi_\theta(a_t\mid s_t)$ incorporates the joint probability of branch selection and node-wise Bernoulli sampling. 
For migration actions, it accounts for the probability of branch selection and target partition selection.

During training, the policy iteratively refines the warm-start partitioning within a bounded horizon $T$, guided by the Cognitive Estimator and the communication simulator. 
During inference, the learned policy deterministically refines the initial partitioning to obtain subregions that are both feasibility-satisfying and communication-efficient.

\section{Prompt Design}
\label{app:prompt_design}

This section presents the prompt templates used by the spatial and temporal agents. 
The spatial agent performs local reasoning and cross-region communication, while the temporal agent ensures temporal validity. 
The prompt templates for the spatial and temporal agents across the four tasks are shown in the following figures.

\begin{figure*}[t]
\centering
\begin{tcolorbox}[
    enhanced,
    width=0.92\textwidth,
    colback=white,
    colframe=black,
    boxrule=0.8pt,
    arc=10pt,
    left=10pt,
    right=10pt,
    top=16pt,
    bottom=10pt,
    overlay={
        \node[
            fill=green!12,
            draw=black,
            rounded corners=4pt,
            line width=0.6pt,
            anchor=west,
            inner xsep=10pt,
            inner ysep=6pt,
            font=\bfseries
        ] at ([xshift=18pt,yshift=-2pt]frame.north west) {Spatial Agent Prompt for Reachability};
    }
]
\scriptsize

\textbf{System Prompt.}

You are an expert in region undirected-graph reachability. 
You will receive \texttt{REGION DATA}. 
\texttt{Region\_edges} is an undirected graph within your region: each pair \texttt{[u,v]} means an undirected edge between nodes $u$ and $v$. 
\texttt{Boundary\_edges} lists boundary connections: each pair \texttt{[b,h]} means a boundary link from your region's boundary node $b$ to a neighboring region's handoff node $h$. 
You MUST follow the TASK and RULES to decide your action.

\vspace{0.5em}
\textbf{REGION DATA.}

\begin{tabular}{@{}ll@{}}
\texttt{Region\_id:}       & \texttt{\{region\_id\}} \\
\texttt{Region\_nodes:}    & \texttt{\{region\_nodes\}} \\
\texttt{Boundary\_nodes:}  & \texttt{\{boundary\_nodes\}} \\
\texttt{Region\_edges:}    & \texttt{\{edges\_str\}} \\
\texttt{Boundary\_edges:}  & \texttt{\{boundary\_str\}} \\
\texttt{Previous\_path:}   & \texttt{\{path\_history\}} \\
\texttt{Current\_node:}    & \texttt{\{current\_node\}} \\
\texttt{Target\_node:}     & \texttt{\{target\_node\}} \\
\end{tabular}

\vspace{0.5em}
\textbf{TASK.}

Choose exactly one action from \texttt{["found", "handoff", "stuck"]}.

\vspace{0.5em}





\textbf{CASE A.}
If \texttt{target\_in\_region == true}, find a valid path from \texttt{Current\_node} to \texttt{Target\_node}. 
If a valid path exists, output \texttt{action="found"} with \texttt{path\_segment}. 
The field \texttt{path\_segment} may contain multiple paths when multiple valid paths exist. 
Otherwise, go to CASE B.

\vspace{0.5em}
\textbf{CASE B.}
If \texttt{target\_in\_region == false} OR no valid path to the target exists, evaluate boundary handoff candidates. 
Let \texttt{degree(x)} be the number of \texttt{Region\_edges} containing $x$. 
If \texttt{degree(Current\_node) == 0} AND \texttt{Current\_node} is in \texttt{Boundary\_nodes}, set \texttt{Boundary\_nodes = [Current\_node]}; otherwise keep \texttt{Boundary\_nodes} unchanged. 
For each boundary node $b$, the agent MUST find a valid path from \texttt{Current\_node} to $b$. 
If such a path exists, then for each boundary edge $e$ in \texttt{Boundary\_edges} where \texttt{e[0] == b}, let \texttt{h = e[1]}. 
If both $b$ and $h$ are not in \texttt{Previous\_path}, create a candidate with \texttt{boundary\_node=b}, \texttt{handoff\_node=h}, and \texttt{path\_to\_node=path}. 
The field \texttt{path\_to\_node} may contain multiple paths when multiple valid paths exist. 
Keep up to \texttt{\{max\_candidates\}} candidates, preferring shorter \texttt{path\_to\_node} and diverse \texttt{handoff\_node}. 
If candidates are not empty, output \texttt{action="handoff"} with candidates.

\vspace{0.5em}
\textbf{OUTPUT JSON.}

\begin{tabular}{@{}l@{}}
\texttt{\{"action":"found", "target\_in\_region":true,}
\quad \texttt{"path\_segment":[...], ...\}} \\

\texttt{\{"action":"handoff", "target\_in\_region":true/false,} 
\quad \texttt{"candidates":[\{"boundary\_node":int,} \\
\quad \texttt{"handoff\_node":int, "path\_to\_node":[...], ...\}]\}} \\

\texttt{\{"action":"stuck", "target\_in\_region":true/false\}}
\end{tabular}

\end{tcolorbox}
\caption{Prompt template for the spatial agent. The agent performs local reachability reasoning and decides whether to return a found path, hand off to another region, or report failure.}
\label{fig:spatial_prompt}
\end{figure*}

\begin{figure*}[t]
\centering
\begin{tcolorbox}[
    enhanced,
    width=0.92\textwidth,
    colback=white,
    colframe=black,
    boxrule=0.8pt,
    arc=10pt,
    left=10pt,
    right=10pt,
    top=16pt,
    bottom=10pt,
    overlay={
        \node[
            fill=red!18,
            draw=black,
            rounded corners=4pt,
            line width=0.6pt,
            anchor=west,
            inner xsep=10pt,
            inner ysep=6pt,
            font=\bfseries
        ] at ([xshift=18pt,yshift=-2pt]frame.north west) {Temporal Agent Prompt for Reachability};
    }
]
\scriptsize

\textbf{System Prompt.}

You are an expert agent specialized in temporal reasoning tasks. 
You will receive a Temporal Definition, Task Instruction, Answer Instruction, and a Question. 
You must solve the temporal task and output the necessary reasoning and the final answer as ``Yes'' or ``No''.

\vspace{0.5em}
\textbf{Temporal Definition.}

A temporal path is a sequence of one or more quadruples 
$(u_i, v_i, s_i, t_i)$, where the undirected edge $(u_i, v_i)$ is added at time $s_i$, and $t_i$ is the last valid time. Equivalently, the deletion happens at time $t_i+1$.
The path is valid iff there exist traversal times $m_1 \leq \cdots \leq m_k$
where equality is allowed, such that for each $i$, $\max(m_{i-1}, s_i) \leq m_i \leq t_i, \text{with } m_0=-1.$

\vspace{0.5em}
\textbf{Validity Check.}

To check validity, greedily assign the earliest possible traversal time for each edge: $m_i = \max(m_{i-1}, s_i)$. If $m_i \leq t_i$ holds for all edges, the path is valid. Remember: any single-quadruple path is valid.

\vspace{0.5em}
\textbf{Task Instruction.}

Your task is to determine whether at least one of the given temporal paths is valid.

\vspace{0.5em}
\textbf{Answer Instruction.}

After \texttt{Answer:}, output only ``Yes'' or ``No''.

\vspace{0.5em}
\textbf{Question.}

Given all reachable paths between node \texttt{\{u\_query\}} and node \texttt{\{v\_query\}}:$\texttt{\{paths\_repr\}}$.
Does there exist a valid temporal path between node \texttt{\{u\_query\}} and node \texttt{\{v\_query\}}?

\vspace{0.5em}
\texttt{Answer:}

\end{tcolorbox}
\caption{Prompt template for the temporal agent. The agent verifies whether candidate paths satisfy temporal validity constraints.}
\label{fig:temporal_prompt}
\end{figure*}

\begin{figure*}[t]
\centering
\begin{tcolorbox}[
    enhanced,
    width=0.92\textwidth,
    colback=white,
    colframe=black,
    boxrule=0.8pt,
    arc=10pt,
    left=10pt,
    right=10pt,
    top=16pt,
    bottom=10pt,
    overlay={
        \node[
            fill=green!12,
            draw=black,
            rounded corners=4pt,
            line width=0.6pt,
            anchor=west,
            inner xsep=10pt,
            inner ysep=6pt,
            font=\bfseries
        ] at ([xshift=18pt,yshift=-2pt]frame.north west) {Spatial Agent Prompt for Temporal Motif Counting};
    }
]
\scriptsize

\textbf{System Prompt.}

You are an expert in undirected graph motif counting. 
\texttt{Region\_edges} is an undirected graph: each pair \texttt{[u, v]} means an undirected edge between nodes $u$ and $v$. 
Do NOT output code, pseudocode, or programming-language snippets. 
Do NOT describe hypothetical execution, e.g., ``after executing the code''. 
Execute the requested steps directly in plain text, following the user's rules exactly. 
Do not replace step-by-step membership checks with shorthand expressions.

\vspace{0.5em}
\textbf{REGION DATA.}

\begin{tabular}{@{}ll@{}}
\texttt{Region\_id:} & \texttt{\{region\_id\}} \\
\texttt{Region\_nodes:} & \texttt{\{region\_nodes\}} \\
\texttt{Region\_edges (undirected):} & \texttt{\{edges\_str\}} \\
\end{tabular}

\vspace{0.5em}
\textbf{MOTIF DEFINITION.}

\texttt{\{motif\_type\}}: A sequence of \texttt{\{l\}} edges, 
$M=\texttt{\{motif\_edge\}}$, involving \texttt{\{k\}} unique nodes.

\vspace{0.5em}
\textbf{TASK.}

Find all \texttt{\{motif\_type\}} using ONLY \texttt{Region\_edges}. 
There is no need to avoid duplicate triangles. 
Output all triangles as they are found.







\vspace{0.5em}
\textbf{OUTPUT JSON.}

\texttt{\{"triangles": [[[u0,u1],[u1,u2],[u2,u0]], ...]\}}.

\end{tcolorbox}
\caption{Prompt template for the spatial agent in the motif recognition task. The agent constructs adjacency sets within a region and enumerates motif instances using only local region edges.}
\label{fig:spatial_prompt_motif}
\end{figure*}

\begin{figure*}[t]
\centering
\begin{tcolorbox}[
    enhanced,
    width=0.92\textwidth,
    colback=white,
    colframe=black,
    boxrule=0.8pt,
    arc=10pt,
    left=10pt,
    right=10pt,
    top=16pt,
    bottom=10pt,
    overlay={
        \node[
            fill=red!18,
            draw=black,
            rounded corners=4pt,
            line width=0.6pt,
            anchor=west,
            inner xsep=10pt,
            inner ysep=6pt,
            font=\bfseries
        ] at ([xshift=18pt,yshift=-2pt]frame.north west) {Temporal Agent Prompt for Temporal Motif Counting};
    }
]
\scriptsize

\textbf{System Prompt.}

You are an expert agent specialized in temporal reasoning tasks. 
Do NOT output code, pseudocode, or programming-language snippets. 
Do NOT describe hypothetical execution, e.g., ``after executing the code''. 
Execute the requested steps directly in plain text, following the user's rules exactly.

\vspace{0.5em}
\textbf{Temporal Definition.}

A temporal motif instance is defined as a sequence of edges
$[[u_0,v_0,t_0], \ldots, [u_l,v_l,t_l]]$,
where timestamps are strictly distinct and $\texttt{time\_span} < \texttt{\{delta\}}$.

\vspace{0.5em}
\textbf{Task.}

Given a list of \texttt{\{motif\_type\}} motif instances, count how many are valid.










\vspace{0.5em}
\textbf{Input Data.}

\begin{tabular}{@{}ll@{}}
\texttt{Motif instances:} \texttt{\{instances\_str\}}. \\
\end{tabular}

\vspace{0.5em}
\textbf{OUTPUT JSON.}

\texttt{\{"Count": count\}}.

\end{tcolorbox}
\caption{Prompt template for the temporal agent in the motif recognition task. The agent verifies temporal validity by checking timestamp distinctness and the temporal span of each candidate motif instance.}
\label{fig:temporal_prompt_motif}
\end{figure*}

\begin{figure*}[t]
\centering
\begin{tcolorbox}[
    enhanced,
    width=0.92\textwidth,
    colback=white,
    colframe=black,
    boxrule=0.8pt,
    arc=10pt,
    left=10pt,
    right=10pt,
    top=16pt,
    bottom=10pt,
    overlay={
        \node[
            fill=green!12,
            draw=black,
            rounded corners=4pt,
            line width=0.6pt,
            anchor=west,
            inner xsep=10pt,
            inner ysep=6pt,
            font=\bfseries
        ] at ([xshift=18pt,yshift=-2pt]frame.north west) {Spatial Agent Prompt for Community Detection};
    }
]
\scriptsize

\textbf{System Prompt.}

You are an expert in community detection using label propagation. 
Perform exactly ONE asynchronous label-update round for \texttt{Region\_nodes}. 
Neighbor relationships come ONLY from \texttt{Edges}. 
\texttt{Edges} is undirected: each pair \texttt{[u, v]} means an undirected edge between nodes $u$ and $v$. 
ASYNCHRONOUS means that nodes are updated in the given \texttt{Region\_nodes} order, and the latest available labels within the same iteration are immediately used.

\vspace{0.5em}
\textbf{REGION DATA.}

\begin{tabular}{@{}p{0.28\linewidth}p{0.62\linewidth}@{}}
\texttt{Region\_id:} & \texttt{\{region\_id\}} \\
\texttt{Iteration:} & \texttt{\{iteration\}} \\
\texttt{Region\_nodes:} & \texttt{\{local\_nodes\}} \\
\texttt{Neighbor\_nodes:} & \texttt{\{neighbor\_nodes\_set\}} \\
\texttt{Edges (Undirected):} & \texttt{\{all\_edges\}} \\
\end{tabular}

\vspace{0.5em}
\textbf{INITIAL LABELS.}

\texttt{All\_node\_labels\_prev} \texttt{(\{node: label\}):}  \texttt{\{all\_label\_map\}} \\

\vspace{0.5em}
\textbf{TASK.}

Do exactly ONE asynchronous update round for \texttt{Region\_nodes} only, and output \texttt{updated\_labels}.






\vspace{0.5em}
\textbf{IMPORTANT.}

When computing \texttt{neighbors(x)}, both endpoints $u$ and $v$ of each edge must be checked to see whether either equals $x$. 
The agent updates \texttt{L[x]} immediately after each node. 
Therefore, when processing the next node in \texttt{Region\_nodes}, if it is a neighbor of a previously updated node, it must use the new label of that neighbor. 
The keys of \texttt{updated\_labels} must match \texttt{Region\_nodes} exactly, with no missing or extra nodes. 
Do NOT output labels for \texttt{Neighbor\_nodes}. 
\texttt{neighbors(x)} MUST contain all and only the neighbors of $x$, with no missing neighbors and no extra nodes. 
If \texttt{current\_label} is in \texttt{tied\_labels}, the agent MUST keep it and must NOT choose the smallest label. 
The output must be a dictionary mapping node to label.

\vspace{0.5em}
\textbf{OUTPUT JSON.}

\texttt{\{"updated\_labels": \{"Region\_nodes": label\}\}}

\end{tcolorbox}
\caption{Prompt template for the spatial agent in the community detection task. The agent performs one asynchronous label-propagation update round within a region using only undirected edge relationships.}
\label{fig:spatial_prompt_community}
\end{figure*}

\begin{figure*}[t]
\centering
\begin{tcolorbox}[
    enhanced,
    width=0.92\textwidth,
    colback=white,
    colframe=black,
    boxrule=0.8pt,
    arc=10pt,
    left=10pt,
    right=10pt,
    top=16pt,
    bottom=10pt,
    overlay={
        \node[
            fill=red!18,
            draw=black,
            rounded corners=4pt,
            line width=0.6pt,
            anchor=west,
            inner xsep=10pt,
            inner ysep=6pt,
            font=\bfseries
        ] at ([xshift=18pt,yshift=-2pt]frame.north west) {Temporal Agent Prompt for Community Detection};
    }
]
\scriptsize

\textbf{System Prompt.}

You are an expert agent specialized in temporal filtering.

\vspace{0.5em}
\textbf{Dynamic Graph Definition.}

A dynamic graph is given as a list of quadruples. 
Each quadruple \texttt{[u, v, t0, t1]} means an undirected edge $(u,v)$ is active from time $t_0$ inclusive to time $t_1$ inclusive. 
That is, the edge exists for all $t$ such that
$t_0 \leq t \leq t_1$.

\vspace{0.5em}
\textbf{Task.}

Given a time window
$[\texttt{W\_start}, \texttt{W\_end}]
=
[\texttt{\{window\_start\}}, \texttt{\{window\_end\}}]$,
select all edges that exist for the ENTIRE window. 
An edge satisfies this condition iff
$t_0 \leq \texttt{W\_start}
\quad \text{AND} \quad
t_1 \geq \texttt{W\_end}$.





\vspace{0.5em}
\textbf{Input.}

\texttt{Dynamic Graph:}  \texttt{\{interval\_str\}} \\

\vspace{0.5em}
\textbf{Output JSON.}

$\texttt{\{"edges": [[u, v], ...]\}}$.

\end{tcolorbox}
\caption{Prompt template for the temporal agent in the Community Detection task. The agent selects edges that remain active throughout the given time window.}
\label{fig:temporal_prompt_filtering}
\end{figure*}

\begin{figure*}[t]
\centering
\begin{tcolorbox}[
    enhanced,
    width=0.92\textwidth,
    colback=white,
    colframe=black,
    boxrule=0.8pt,
    arc=10pt,
    left=10pt,
    right=10pt,
    top=16pt,
    bottom=10pt,
    overlay={
        \node[
            fill=green!12,
            draw=black,
            rounded corners=4pt,
            line width=0.6pt,
            anchor=west,
            inner xsep=10pt,
            inner ysep=6pt,
            font=\bfseries
        ] at ([xshift=18pt,yshift=-2pt]frame.north west) {Spatial Agent Prompt for Connected Components};
    }
]
\scriptsize

\textbf{System Prompt.}

You are an expert in undirected graph connected-components counting. \\

\vspace{0.5em}
\textbf{REGION DATA.}

\begin{tabular}{@{}p{0.28\linewidth}p{0.62\linewidth}@{}}
\texttt{Region\_id:} & \texttt{\{region\_id\}} \\
\texttt{Region\_nodes:} & \texttt{\{region\_nodes\}} \\
\texttt{Boundary\_nodes:} & \texttt{\{boundary\_str\}} \\
\texttt{Region\_edges (undirected):} & \texttt{\{edges\_str\}} \\
\end{tabular}

\vspace{0.5em}
\textbf{Connected Component Definition.}

In an undirected graph, a connected component is a maximal subgraph in which every pair of nodes is connected by some path.

\vspace{0.5em}
\textbf{TASK.}

Compute all connected components within the region using ONLY \texttt{Region\_edges}, then output:
1) the number of connected components; 
2) for each boundary node, the \texttt{component\_id} it belongs to.








\vspace{0.5em}
\textbf{Critical.}

Components must cover all \texttt{Region\_nodes} exactly once, with full cover and no overlap. For nodes that form singleton components, verify that they indeed have no incident edges; otherwise, union them with the corresponding component.  \texttt{local\_cc\_count == len(components)}. Keys of \texttt{boundary\_cc\_map} must exactly match \texttt{Boundary\_nodes}. And for each boundary node, \texttt{boundary\_cc\_map[node]} must be the minimum node id of its component.

\vspace{0.5em}
\textbf{OUTPUT JSON.}

$\texttt{\{"local\_cc\_count": int, "boundary\_cc\_map": \{"boundary\_node": component\_id\}\}}$

\end{tcolorbox}
\caption{Prompt template for the spatial agent in the connected components task. The agent computes local connected components within a region and maps each boundary node to its component identifier.}
\label{fig:spatial_prompt_connected_components}
\end{figure*}

\begin{figure*}[t]
\centering
\begin{tcolorbox}[
    enhanced,
    width=0.92\textwidth,
    colback=white,
    colframe=black,
    boxrule=0.8pt,
    arc=10pt,
    left=10pt,
    right=10pt,
    top=16pt,
    bottom=10pt,
    overlay={
        \node[
            fill=red!18,
            draw=black,
            rounded corners=4pt,
            line width=0.6pt,
            anchor=west,
            inner xsep=10pt,
            inner ysep=6pt,
            font=\bfseries
        ] at ([xshift=18pt,yshift=-2pt]frame.north west) {Temporal Agent Prompt for Connected Components};
    }
]
\scriptsize

\textbf{System Prompt.}

You are an expert agent specialized in temporal filtering.

\vspace{0.5em}
\textbf{Dynamic Graph Definition.}

A dynamic graph is given as a list of quadruples. 
Each quadruple $[u, v, t_0, t_1]$ means that an undirected edge $(u,v)$ is active from time $t_0$ inclusive to time $t_1$ inclusive. 
That is, the edge exists for all $t$ such that
$t_0 \leq t \leq t_1$.

\vspace{0.5em}
\textbf{Task.}

Given a time window
$[\texttt{W\_start}, \texttt{W\_end}]
=
[\texttt{\{window\_start\}}, \texttt{\{window\_end\}}]$,
select all edges that exist for the ENTIRE window. 
An edge should be selected iff it satisfies:
$(t_0 \leq \texttt{W\_start})$
AND
$(t_1 \geq \texttt{W\_end})$.





\vspace{0.5em}
\textbf{Input.}

\texttt{Dynamic Graph:} \texttt{\{interval\_str\}} \\

\vspace{0.5em}
\textbf{Output JSON.}

$\texttt{\{"edges": [[u, v], ...]\}}$

\end{tcolorbox}
\caption{Prompt template for the temporal agent in the connected components task. The agent filters dynamic edges and returns the edges that remain active throughout the given time window.}
\label{fig:temporal_prompt_connected_components}
\end{figure*}



\end{document}